\documentclass[pmlr,twocolumn,10pt,letterpaper]{jmlr} %

\usepackage{booktabs}
\usepackage{siunitx}
\usepackage{pdfpages}

\usepackage[switch]{lineno}

\theorembodyfont{\upshape}
\theoremheaderfont{\scshape}
\theorempostheader{:}
\theoremsep{\newline}

\usepackage{multirow}
\usepackage{stfloats}
\usepackage[htt]{hyphenat}

\usepackage{xcolor}
\usepackage{color}
\usepackage{soul}

\newenvironment{ttblock}{\ttfamily}{}
\newcommand{\highlightcolor}[2]{%
\sethlcolor{#1}%
\hl{#2}%
}

\definecolor{oc-blue-0}{HTML}{E7F5FF}
\definecolor{oc-blue-7}{HTML}{1C7ED6}
\definecolor{oc-blue-8}{HTML}{1971C2}
\definecolor{oc-blue-9}{HTML}{1864AB}
\definecolor{oc-lime-0}{HTML}{F4FCE3}
\definecolor{oc-lime-8}{HTML}{66A80F}
\definecolor{oc-lime-9}{HTML}{5C940D}
\definecolor{oc-green-0}{HTML}{EBFBEE}
\definecolor{oc-green-8}{HTML}{2F9E44}
\definecolor{oc-green-9}{HTML}{2B8A3E}
\definecolor{oc-orange-0}{HTML}{FFF4E6}
\definecolor{oc-orange-8}{HTML}{E8590C}
\definecolor{oc-orange-9}{HTML}{D9480F}
\newcommand{\styledtext}[3]{\textcolor{#1}{\highlightcolor{#2}{#3}}}
\newcommand{\prompttemplate}[1]{\styledtext{oc-blue-9}{oc-blue-0}{#1}}
\newcommand{\promptgeneration}[1]{\styledtext{oc-green-9}{oc-green-0}{#1}}

\jmlrvolume{248}
\jmlryear{2024}
\jmlrworkshop{Conference on Health, Inference, and Learning (CHIL) 2024} %
\firstpageno{339}

\title[A Data-Centric Approach To Generate Faithful and High Quality Patient Summaries with LLMs]{A Data-Centric Approach To Generate Faithful and High Quality Patient Summaries with Large Language Models}

\author{%
\renewcommand{\thefootnote}{\arabic{footnote}}\Name{Stefan Hegselmann}\footnotemark[1]\Email{stefan.hegselmann@uni-muenster.de}\\
\Name{Shannon Zejiang Shen}\footnotemark[2] \Email{zjshen@mit.edu}\\
\Name{Florian Gierse}\footnotemark[1] \Email{flogierse@uni-muenster.de}\\
\Name{Monica Agrawal}\footnotemark[3] \Email{monica.agrawal@duke.edu}\\
\Name{David Sontag}\footnotemark[2] \Email{dsontag@csail.mit.edu}\\
\Name{Xiaoyi Jiang}\footnotemark[1] \Email{xjiang@uni-muenster.de}\\ %
\addr \footnotemark[1]University of Münster, Germany \\
\footnotemark[2]MIT CSAIL, Cambridge, MA, USA \\
\footnotemark[3]Duke University, Durham, NC, USA
}

\begin{document}

\maketitle

\begin{abstract}
Patients often face difficulties in understanding their hospitalizations, 
while healthcare workers have limited resources to provide explanations.
In this work, we investigate the potential of large language models to generate patient summaries based on doctors' notes and study the effect of training data on the faithfulness and quality of the generated summaries.
To this end, we release (i) a rigorous labeling protocol for errors in medical texts and (ii) a publicly available dataset of annotated hallucinations in 100 doctor-written and 100 generated summaries.
We show that fine-tuning on hallucination-free data effectively reduces hallucinations from 2.60 to 1.55 per summary for Llama 2, while preserving relevant information.
We observe a similar effect on GPT-4 (0.70 to 0.40), when the few-shot examples are hallucination-free.
We also conduct a qualitative evaluation using hallucination-free and improved training data.
We find that common quantitative metrics do not correlate well with faithfulness and quality.
Finally, we test GPT-4 for automatic hallucination detection, which clearly outperforms common baselines.
\end{abstract}

\paragraph*{Data and Code Availability}
We use \texttt{MIMIC-IV-Note} \citep{johnson_mimic-iv-note_2023, goldberger_physiobank_2000}.
We create a dataset of discharge instructions (\texttt{MIMIC-IV-Note-Ext-DI}) and hallucination annotations for 100 doctor-written and 100 generated patient summaries (\texttt{Hallucinations-\{MIMIC,Generated\}-DI}). Our data and code: \url{https://doi.org/10.13026/m6hf-dq94} and \url{https://github.com/stefanhgm/patient_summaries_with_llms}.

\paragraph*{Institutional Review Board (IRB)}
Our work did not require IRB approval.
The two clinical annotators are authors of this paper, and both had credentialed access to the \texttt{MIMIC-IV-Note} dataset.

\section{Introduction}
\label{sec:intro}
\begin{figure*}[t!]
  \centering
  \includegraphics[width=1.0\textwidth]{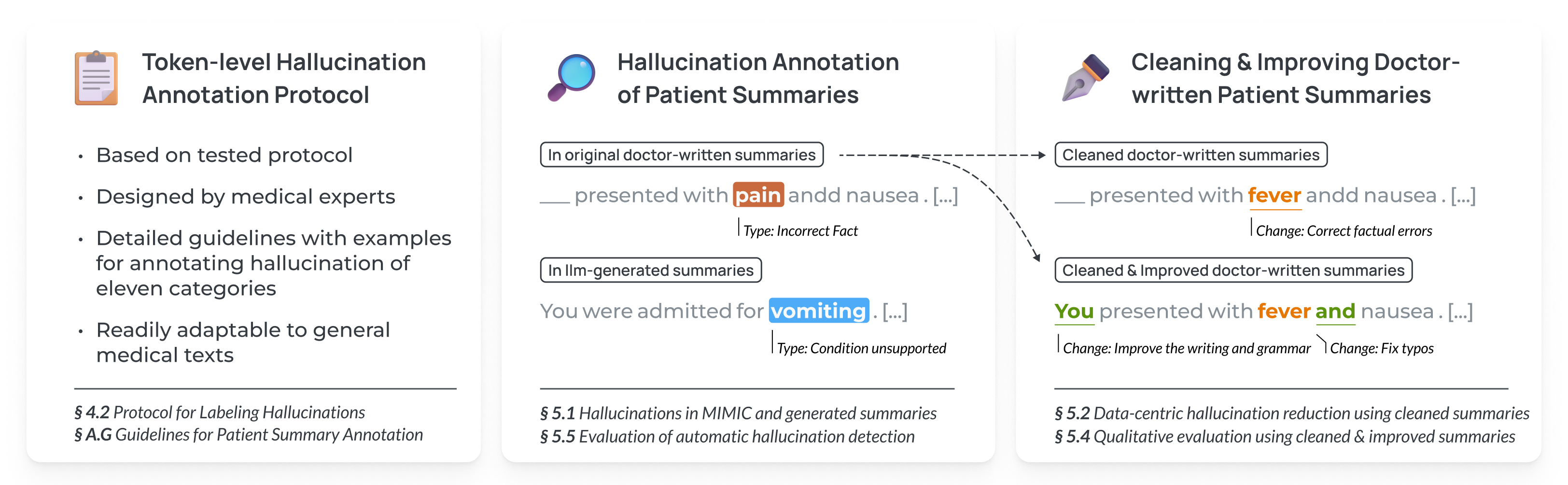}
  \caption{We developed a protocol for annotating hallucinations in medical text. Following this protocol, two medical experts labeled hallucinations in 100 doctor-written (\texttt{Hallucinations-MIMIC-DI}) and 100 LLM-generated patient summaries (\texttt{Hallucinations-Generated-DI}). We used the labeled hallucinations in the doctor-written summaries to derive two additional datasets by replacing or removing hallucinations (\texttt{Cleaned}) and by further improving the language (\texttt{Cleaned \& Improved}). We used these two datasets for our data-centric hallucination reduction and qualitative experiments.}

  \label{fig:overview}
\end{figure*}

Many patients do not understand the events that occurred during their hospitalization and the subsequent actions they need to take \citep{kebede_patients_2014}.
For instance, \cite{horwitz_quality_2013} performed post-discharge interviews and found that only 59.6\% of the patients were able to accurately describe their admission diagnosis and 43.9\% could fully describe their scheduled follow-up appointments.
Improved discharge communication is associated with lower hospital readmission rates and higher adherence to treatment regiments \citep{becker_interventions_2021}.
A potential intervention to improve patient comprehension could be patient-oriented summaries that describe all relevant facts in layperson language \citep{federman_challenges_2018}.
However, writing high-quality patient summaries is a difficult and time-consuming task \citep{mueller_readability_2015}, and healthcare workers already face high workloads \citep{phillips_relationships_2020,watson_impact_2019}.

Large language models (LLMs) have demonstrated strong capabilities on many natural language tasks including medical summarization \citep{van_veen_adapted_2024}.
However, LLMs are prone to generating unsupported or erroneous facts also referred to as hallucinations \citep{maynez_faithfulness_2020}.
In healthcare, this issue is further aggravated by the fragmented nature of healthcare data, as datasets often do not perfectly mimic the data available at the point of care.
For example, radiologists often compare to previous images in their reports, even though many medical imaging datasets do not have the historical images available.
Similarly, datasets for medical summarization may not include the full patient history to accompany the written summarization.
As a result, there may be references in the summary to information that is not supported by the available context or patient history.
Training or fine-tuning on this data replicates these artifacts leading to ``hallucinations''.
Several techniques for preventing hallucinations have been studied \citep{huang_factual_2023}. %
However, hallucinations can vary highly in complexity, escaping automatic detection and making careful human annotation necessary \citep{thomson_evaluating_2023,moramarco_human_2022}.

Recent work has emphasized the sample efficiency of LLMs.
\cite{zhou_lima_2023} showed that 1,000 carefully selected fine-tuning examples sufficed for the successful alignment of Llama 2 \citep{touvron_llama_2023}.
An additional set of 30 examples was found to enable multi-turn dialogues \citep{zhou_lima_2023}.
Also, GPT-4 achieved state-of-the-art performance on many benchmarks using at most 25 in-context examples \citep{openai_gpt-4_2023}.
These findings open up opportunities for data-centric approaches that leverage data curated by human domain experts.
In this study, we manually create 100 training examples without hallucinations and 100 training examples with improved quality to study their effect on the faithfulness and quality of LLMs.
Our main contributions are:

\begin{enumerate}
    \item We introduce a rigorous annotation protocol for token-level errors in medical texts.
    \item We release two datasets of 100 doctor-written (MIMIC) and 100 generated patient summaries with hallucinations labeled by two trained medical experts requiring 70 hours per annotator.
    \item For patient summaries, we demonstrate that fine-tuning on data with manually removed hallucinations can effectively reduce the hallucinations in LLMs while preserving key information.
    
    \item We evaluate GPT-4 for automatic hallucination detection using our datasets, emphasizing its suitability for identifying unsupported evidence.
    
\end{enumerate}

\section{Related Work}
\label{sec:related_work}
Given the large burden of and repetition within clinical documentation, automated clinical summarization is of great interest across healthcare; generation targets include discharge summaries, radiology impression sections, patient-facing instructions, and problem lists \citep{van_veen_adapted_2024, adams_learning_2022}. 
Consequently, there has been significant research in the past few years exploring how the advent of LLMs can be leveraged to automate clinical summarization.
Approaches for control have included both zero- and few-shot prompting \citep{van_veen_adapted_2024} to fine-tuning \citep{adams_learning_2022, adams_meta-evaluation_2023, moramarco_human_2022, cai_generation_2022, van_veen_radadapt_2023}.
Results have been promising; for example, \citet{van_veen_adapted_2024} found that GPT-4 is often preferred over human-generated summaries in terms of correctness and completeness. 

However, hallucinations -- the generation of facts not grounded in the original text --- are a recurring concern in text summarization, both across natural language processing (NLP) in general and specifically in the clinical domain \citep{zhang_optimizing_2020, kryscinski_evaluating_2020, adams_learning_2022, cai_generation_2022, xie_factreranker_2023}.
Given the ubiquity of this issue, several taxonomies have sprung up in the wider NLP community for producing fine-grained annotations of the accuracy of generated text \citep{thomson_gold_2020, mishra_fine-grained_2024}.
In the clinical domain, there have been similar efforts to annotate hallucinations in generated text, but this work has often occurred on proprietary data, with coarse buckets, or without the release of annotations \citep{adams_meta-evaluation_2023, cai_generation_2022, van_veen_adapted_2024}.
The closest public release was a set of edits clinicians made to summaries auto-generated from transcripts, but this did not target hallucinations specifically \citep{moramarco_human_2022}. 
To our knowledge, this work is the first to release a dataset highlighting hallucinations in clinical summarization.
These are particularly important because automated metrics of summarization quality, while fast to compute, often do not correlate with manually evaluated performance, a finding also replicated in this work \citep{van_veen_adapted_2024}.

There have been three major strategies to mitigate the effect of hallucinations in auto-generated summaries.
The first is to conduct post-hoc detection of hallucinations. This has largely involved learning a model of hallucinations from synthetically generated data \citep{cai_generation_2022, zhou_detecting_2021, kryscinski_evaluating_2020} or identifying normalized concepts (e.g., UMLS concepts) that appear in the summary, but not the source \citep{nan_entity-level_2021, adams_learning_2022}.
The second strategy is to minimize hallucinations by changing the underlying generation model, e.g., by grounding in Wikipedia, or first converting the source context as triples as an intermediary before generation \citep{zhang_optimizing_2020, tian_fine-tuning_2023, semnani_wikichat_2023, aralikatte_focus_2021, cao_faithful_2018, cao_cliff_2021}.
The final strategy is to improve the quality of the training data itself, either by removing poor training examples \citep{nan_entity-level_2021} or by improving the quality of the training examples.
For example, to avoid decreasing sample size, \citet{adams_learning_2022} revised the reference text via contrastive learning on synthetic data.
In this work, we show that for sample-efficient LLMs, it is sufficient to revise only a small number of training examples that are feasible even manually.

\section{Datasets}
\label{sec:dataset}
\subsection{MIMIC Discharge Instructions Datasets}

\begin{table*}[h!] %
\setlength{\tabcolsep}{2.9pt}
\small
\begin{tabular}{@{}lll@{}}
\toprule
\textbf{Dataset} & \textbf{Size} & \textbf{Description} \\ 
\midrule
\multicolumn{3}{@{}l}{MIMIC Discharge Instructions Datasets} \\
\midrule
\texttt{MIMIC-IV-Note-Ext-DI}          & 100,175 & \multirow{2}{*}{\begin{tabular}[c]{@{}l@{}}Summarization dataset derived from \texttt{MIMIC-IV-Note} with the section\\Discharge Instructions as summary and all prior notes as context\end{tabular}} \\
                                   &     &  \\
\texttt{MIMIC-IV-Note-Ext-DI-BHC}      & 100,175 & \texttt{MIMIC-IV-Note-Ext-DI} with the Brief Hospital Course as context \\
\texttt{MIMIC-IV-Note-Ext-DI-BHC-Anno} & 26,178 & \multirow{2}{*}{\begin{tabular}[c]{@{}l@{}}Subset of \texttt{MIMIC-IV-Note-Ext-DI-BHC} with contexts $\leq$ 4,000 characters\\ and summaries $\geq$ 600 characters to facilitate human annotation\end{tabular}} \\
                                   &     &  \\
\midrule
\multicolumn{3}{@{}l}{Hallucination Datasets Annotated by Two Medical Experts} \\
\midrule
\texttt{Hallucinations-MIMIC-DI}     & 100  & Random examples from \texttt{MIMIC-IV-Note-Ext-DI-BHC-Anno} \\
\texttt{Hallucinations-Generated-DI} & 100  & \multirow{2}{*}{\begin{tabular}[c]{@{}l@{}}20 random contexts from \texttt{M.-IV-Note-Ext-DI-BHC-Anno} and summaries \\ generated with five models during hallucination-reduction experiments\end{tabular}} \\
                                     &     &  \\
\midrule
\multicolumn{3}{@{}l}{Derived Datasets from \texttt{Hallucinations-MIMIC-DI}} \\
\midrule
\texttt{Original}             & 100 & Context-summary pairs from \texttt{Hallucinations-MIMIC-DI}\\
\texttt{Cleaned}              & 100 & \texttt{Original} with labeled
hallucinations manually removed or replaced\\
\texttt{Cleaned \& Improved}  & 100 & \texttt{Cleaned} with mistakes and artifacts removed or corrected \\
\bottomrule
\end{tabular}
\caption{Overview of all datasets used in this work. All datasets are publicly available on PhysioNet.
}
\label{tab:datasets}
\end{table*}

For our experiments, we created a summarization dataset with clinical notes as context and patient summaries written by doctors as targets.
We used the MIMIC-IV-Note v2.2 database, which includes 331,793 deidentified clinical notes from 145,915 patients admitted to Beth Israel Deaconess Medical Center in Boston, MA, USA.
Each note consists of various sections that describe a patient's hospital course.
We selected the \textit{Discharge Instructions} (DI) as summaries and the \textit{Brief Hospital Course} (BHC) as contexts since the BHCs contain the most relevant information for medical professionals (see Figure \ref{fig:labelling_example}).
We chose this shorter context to reduce the effort for the human annotators and to better fit it into the models' context windows.
Many DIs contained irrelevant artifacts that could distort the downstream analysis.
For instance, they consisted of static templates or started with a personal salutation.
Hence, we designed a preprocessing pipeline selecting and cleaning 100,175 of the original 331,793 MIMIC-IV-Note examples (see Appendix \ref{apd:data_preprocessing}).
The resulting dataset is named \texttt{MIMIC-IV-Note-Ext-DI-BHC} (see Table \ref{tab:datasets}).
We also release a version of the dataset with a longer context using all note sections before the DI, including the BHC (\texttt{MIMIC-IV-Note-Ext-DI}).
Details for both summarization datasets can be found in Table \ref{tab:dataset_overview_full} in the Appendix.
To further facilitate human annotation, we considered a subset of the data with context lengths of at most 4,000 characters and summary lengths of at least 600 characters.
This was done to reduce the amount of context to take into account for the annotators and 
to increase the information in the summaries.
The resulting subset contained 26,178 entries (\texttt{MIMIC-IV-Note-Ext-DI-BHC-Anno}) and we used it to sample examples for human annotation.

\subsection{Hallucination Datasets}
\label{sec:dataset_hallucination_labelling}

We collected hallucination labels for 100 doctor-written and 100 LLM-generated patient summaries.
The doctor-written summaries were selected at random from \texttt{MIMIC-IV-Note-Ext-DI-BHC-Anno}.
For the LLM-generated summaries, we selected 20 held-out contexts from \texttt{MIMIC-IV-Note-Ext-DI-BHC-Anno} and used five models, which had been trained for the data-centric hallucination reduction experiments, to generate the summaries.
Two medical experts labeled hallucinations in both datasets, following our annotation protocol.
This resulted in two datasets: \texttt{Hallucinations-MIMIC-DI} for the doctor-written summaries and \texttt{Hallucinations-Generated-DI} for the LLM-generated summaries (see Table \ref{tab:datasets}).

\subsection{Derived Datasets}

We define \texttt{Original} as a synonym for the data in \texttt{Hallucinations-MIMIC-DI} because it contains the original 100 doctor-written summaries that may include unsupported facts or errors, referred to as hallucinations.
Based on the human labels of these hallucinations, we derived two additional datasets from \texttt{Hallucinations-MIMIC-DI} (see Table \ref{tab:datasets}).
The \texttt{Cleaned} dataset contains the same patient summaries with annotated hallucinations manually replaced or removed.
We used the \texttt{Original} and \texttt{Cleaned} datasets to test the data-centric hallucination reduction approach (see Section \ref{sec:methods_hallucination_reduction}).
For the \texttt{Cleaned \& Improved} data, we further manually corrected mistakes and artifacts in the summaries.
This dataset was used to fine-tune or prompt the models for our qualitative evaluation (see Section \ref{sec:methods_qualitative_evaluation}).

\section{Methods}
\label{sec:methods}
\subsection{Generation of Patient Summaries}
In this work, we generated patient summaries given the BHC as context.
Formally, we have a set of contexts $\mathcal{C} = \{C_1, ..., C_n\}$ for which we predict the summaries $\mathcal{S} = \{S_1, ..., S_n\}$. 
Note, however, that there is an additional translation step that simplifies the context into layperson language \citep{weng_unsupervised_2019}.
On average, the context $C$ was 552.0 words long, and the summary $S$ was 113.2 words long (see Table \ref{tab:dataset_overview_full} in the Appendix).

We included Llama 2 and GPT-4 in the data-centric hallucination reduction experiments since they are commonly used models that allow for sample-efficient alignment \citep{zhou_lima_2023, openai_gpt-4_2023}.
Since this experiment required expensive manual annotations of hallucinations, we could not test additional models.
For the quantitative and qualitative evaluation, we also included the Longformer Encoder-Decoder (LED), which has shown good performance in medical summarization \citep{cai_generation_2022, adams_learning_2022}.
Further details on parameter tuning for LED and Llama 2, and prompt tuning for GPT-4, are given in Appendices \ref{apd:parameter_tuning} and \ref{apd:prompt_tuning_gpt}.

\begin{itemize}
    \topsep0.1em 
    \itemsep0.1em 
    \item \textbf{LED}: The Longformer Encoder-Decoder was used as a baseline model \citep{beltagy_longformer_2020}.\footnote{Huggingface models \texttt{allenai/led-\{base/large\}-16384}}
    The LED model was initialized by BART \citep{lewis_bart_2020} and can process 16K tokens.
    BART can only handle 1K tokens, which was insufficient.
    For training, we used full fine-tuning on 80\% of \texttt{MIMIC-IV-Note-Ext-DI-BHC}(\texttt{-Anno}).
    The datasets with 100 examples were to small.
    \item \textbf{Llama}: Llama 2 \citep{touvron_llama_2023} has shown promising performance on clinical text summarization \citep{van_veen_adapted_2024}, and we used the versions with 7B and 70B parameters.\footnote{Huggingface models \texttt{meta-llama/Llama-2-\{7,70\}b-hf}}
    We always used 100 training examples for parameter-efficient fine-tuning with LoRA \citep{hu_lora_2021} and loaded the models in 8-bit to reduce the memory usage.
    Due to training time constraints, we did not train with all examples.
    \item \textbf{GPT-4}: GPT-4 \citep{openai_gpt-4_2023} represents the state of the art in clinical summarization \citep{van_veen_adapted_2024}.
    We accessed the model via the Azure OpenAI service with opt-out for human review of the data to ensure data privacy.
    We tested the model with 5 in-context examples (5-shot) or no examples (0-shot).
\end{itemize}

\subsection{Protocol for Labeling Hallucinations}
\label{sec:methods_labelling_hallucinations}

\begin{figure*}[ht!]
  \centering
  \includegraphics[width=1\textwidth]{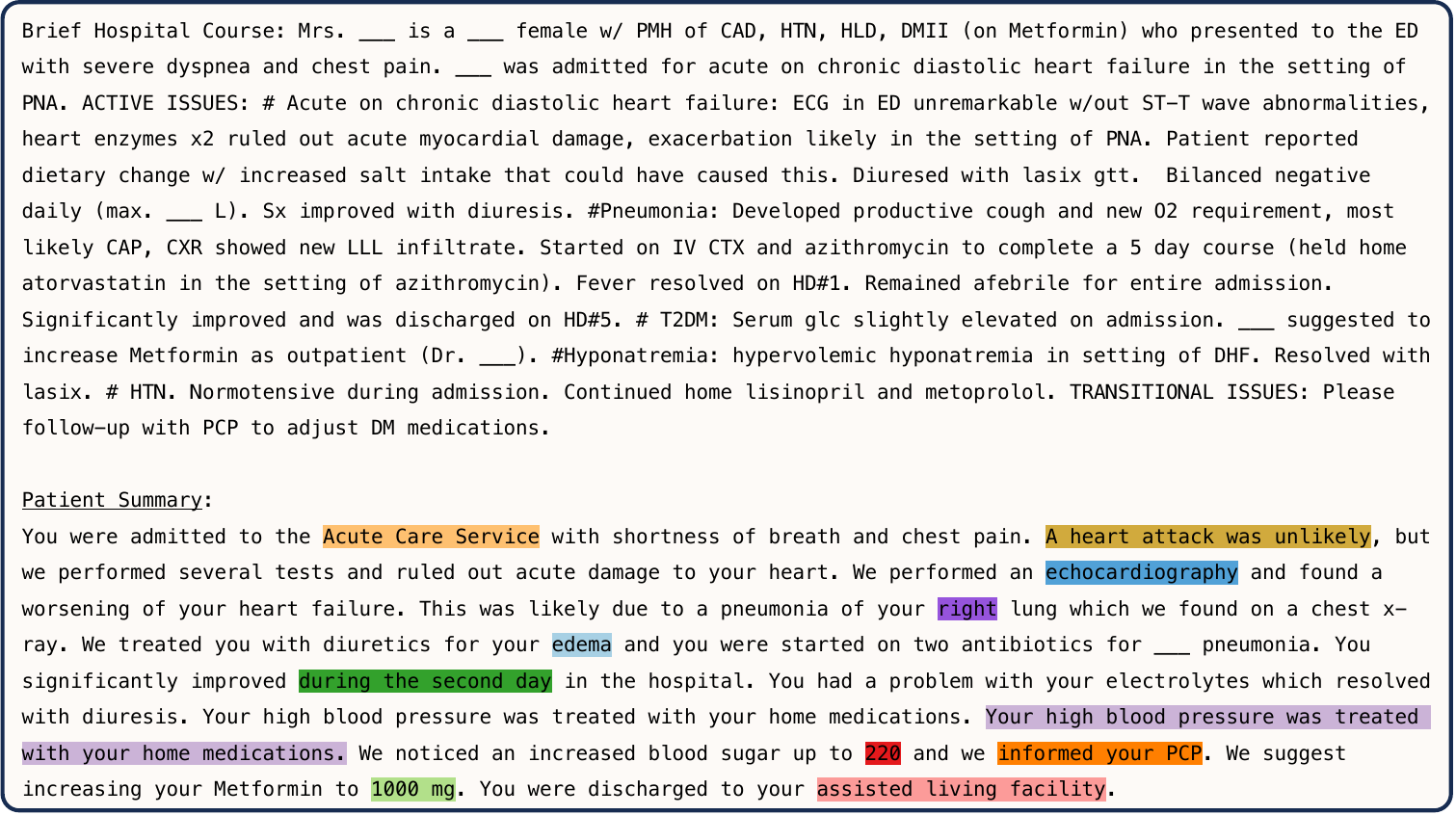}
  \hspace*{1ex}\includegraphics[width=16.2cm]{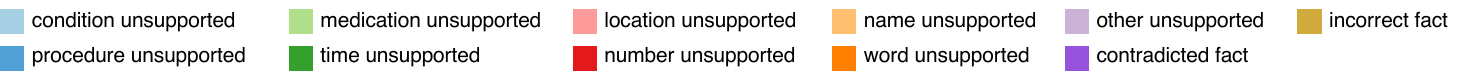}
  \caption{A synthetic MIMIC example labeled with the developed annotation protocol for hallucinations. The protocol was adapted from \cite{thomson_gold_2020} and we used eleven different  labels.}
  \label{fig:labelling_example}
\end{figure*}

We developed a protocol for labeling token-level errors in medical texts based on \citep{thomson_gold_2020,thomson_generation_2021}.\footnote{See the Word document on GitHub: \url{https://github.com/ehudreiter/accuracySharedTask/blob/main/example_exercise/Example_Annotation_Exercise.docx}}
Our main focus was to annotate hallucinations in patient summaries. However, we believe that with slight modifications, it could be applicable to other medical scenarios.
We distinguished between unsupported, contradicted, and incorrect facts.
Since most hallucinations were unsupported facts, we further distinguished them into nine subcategories (see Figure \ref{fig:labelling_example}).
\cite{thomson_gold_2020} considered facts from all sources as given; that is, annotators could also use information from the internet to check facts in the summary.
In contrast to that, we treated the context (BHC) as the only ground truth about the patient.
We chose this approach to reduce the labeling burden, as annotators could not be expected to review all notes and structured information of a patient.
However, since patient summaries contain not only patient-specific information, we did allow \textit{general medical knowledge} and \textit{advice} even if not explicitly provided in the context (e.g., ``Please take your medications as prescribed'').
Clarifications that arose during the annotation process were added to the protocol.
The final labeling protocol can be found in Appendix \ref{apd:protocol}.

\subsection{Hallucinations in MIMIC and Generated Patient Summaries}
\label{sec:methods_hallucination_mimic_generated}

We used our developed protocol to annotate hallucinations in 100 doctor-written MIMIC summaries (\texttt{Hallucinations-MIMIC-DI}) and 100 LLM-generated summaries (\texttt{Hallucinations-Generated-DI}).
It is important to note that ``hallucinations'' in doctor-written summaries are common in healthcare practice and usually should not be regarded as errors.
Doctors may include information in the summary that was never documented, that was documented outside the considered context (in our case, only the BHC), or that was altered just prior to discharge.
In this work, we considered the MIMIC data from a machine learning perspective and analyzed the effect of hallucinations in the training data on LLMs.

The labeling was carried out by two German medical students in their sixth year.
They had completed their second state examination (USMLE Step 2 equivalent) and were working in the hospital.
We utilized MedTator for annotation \citep{he_medtator_2022}.
For annotator training, we used twelve examples.
Two examples were used to familiarize with the task and two times five examples were labeled separately and discussed for training.
For the final labeling, the annotators worked independently and reached a consensus through discussion.
To quantify the variability of the annotation, we determined manual and automatic statistics for agreement (see Appendix \ref{apd:annotation_statistics}).

\subsection{Data-Centric Hallucination Reduction}
\label{sec:methods_hallucination_reduction}

We evaluated whether manually removing hallucinations from the training data can reduce hallucinations of LLMs.
To this end, we tested Llama 70B fine-tuned on 100 examples and GPT-4 5-shot prompted with 5 random examples from the \texttt{Original} and the hallucination-free \texttt{Cleaned} data.
We also included GPT-4 0-shot, which did not require any training data.
Figure \ref{fig:hallucination_examples} contains examples for all five models.
Due to limited manual annotation capabilities, no further models were included in this analysis.
To determine the number of generated hallucinations, we chose 20 random contexts from \texttt{MIMIC-IV-Note-Ext-DI-BHC-Anno} and generated summaries with each of the five models.
We annotated the resulting 100 summaries with our protocol yielding the hallucination dataset \texttt{Hallucinations-Generated-DI}.
We also determined the amount of missing key facts and medical jargon using the annotation procedures of the qualitative evaluation (see Appendix \ref{apd:qualitative_evaluation}).
The summaries for each context were permuted during the annotation process to prevent model identification.

\begin{table*}[ht!]
\small
\setlength{\tabcolsep}{7.2pt}
\begin{tabular}{lcccc}
\toprule
    & \multicolumn{4}{c}{\textbf{Mean (SD)}} \\
\textbf{Model (training data)}                    & Hallucinations$\downarrow$ & Missing Key Facts$\downarrow$ & Medical Jargon$\downarrow$ & Length (Words) \\ 
\midrule
Llama 70B (100 \texttt{Original} ex.) & $2.60$ ($2.39$) & $3.77$ ($1.33$) & $1.05$ ($0.84$) & $97.90$ ($36.73$) \\
Llama 70B (100 \texttt{Cleaned} ex.) & $1.55$ ($1.99$) & $3.73$ ($1.45$) & $1.68$ ($1.23$) & $96.20$ ($31.82$) \\
\midrule
GPT-4 5-shot (5 \texttt{Original} ex.) & $0.70$ ($0.86$) & $0.93$ ($0.80$) & $1.07$ ($0.99$) & $151.10$ ($19.42$) \\
GPT-4 5-shot (5 \texttt{Cleaned} ex.) & $0.40$ ($0.75$) & $0.97$ ($0.80$) & $1.25$ ($1.18$) & $158.80$ ($23.27$) \\
\midrule
GPT-4 0-shot (none) & $0.45$ ($0.60$) & $0.82$ ($0.61$) & $0.70$ ($1.03$) & $165.05$ ($22.75$) \\
\bottomrule
\end{tabular}
\caption{Results for data-centric hallucination-reduction showing mean number of hallucinations, missing key facts, medical jargon, and words generated by Llama 70B and GPT-4 5-shot trained or prompted on \texttt{Original} versus \texttt{Cleaned} summaries (hallucinations removed) and GPT-4 0-shot. Training Llama using hallucination-free summaries shows a strong hallucination reduction while keeping key facts.}
\label{tab:hallucination-statistics}
\end{table*}
\begin{table*}[ht!]
\small
\setlength{\tabcolsep}{2.5pt}
\setlength\doublerulesep{0.5pt} 
\begin{tabular}{lcccccccccccc}
\toprule
\textbf{Dataset /}    & \multicolumn{12}{c}{\textbf{Number of Annotated Hallucinations per Type}} \\
\hspace{0.2cm} \textbf{Model (training data)} & \small{cond.} & \small{proc.} & \small{medic.} & \small{time} & \small{loc.} & \small{numb.} & \small{name} & \small{word} & \small{other} & \small{contrad.} & \small{incorr.} & \small{Total} \\
\midrule
\texttt{Hallucinations-MIMIC-DI} & $52$ & $19$ & $34$ & $35$ & $29$ & $7$ & $18$ & $76$ & $1$ & $15$ & $0$ & $286$  \\
\midrule
\texttt{Hallucinations-Generated-DI} & $27$ & $4$ & $10$ & $2$ & $12$ & $3$ & $5$ & $44$ & $0$ & $7$ & $0$ & $114$  \\
\midrule
\hspace{0.2cm} Llama 70B (100 \texttt{Original} ex.) & $16$ & $2$ & $9$ & $1$ & $4$ & $3$ & $3$ & $11$ & $0$ & $3$ & $0$ & $52$  \\
\hspace{0.2cm} Llama 70B (100 \texttt{Cleaned} ex.) & $7$ & $2$ & $1$ & $1$ & $5$ & $0$ & $1$ & $10$ & $0$ & $4$ & $0$ & $31$  \\
\cmidrule(l{1em}){1-13}
\hspace{0.2cm} GPT-4 5-shot (5 \texttt{Original} ex.) & $2$ & $0$ & $0$ & $0$ & $1$ & $0$ & $1$ & $10$ & $0$ & $0$ & $0$ & $14$  \\
\hspace{0.2cm} GPT-4 5-shot (5 \texttt{Cleaned} ex.) & $1$ & $0$ & $0$ & $0$ & $2$ & $0$ & $0$ & $5$ & $0$ & $0$ & $0$ & $8$  \\
\cmidrule(l{1em}){1-13}
\hspace{0.2cm} GPT-4 0-shot (none) & $1$ & $0$ & $0$ & $0$ & $0$ & $0$ & $0$ & $8$ & $0$ & $0$ & $0$ & $9$  \\
\bottomrule
\end{tabular}
\caption{The first row shows the labeling results for 100 MIMIC summaries for which we found 286 hallucinations. The subsequent rows present the category breakdown for different types of hallucinations annotated in 20 summaries. It corresponds to the data in the \textit{Hallucinations$\downarrow$} column in Table \ref{tab:hallucination-statistics}.}
\label{tab:label-statistics-mimic-models}
\end{table*}

\subsection{Quantitative Evaluation}

We evaluated the performance of all models to compare them to existing work.
We used the \texttt{MIMIC-IV-Note-Ext-DI-BHC} dataset, which contains 100,175 context-summary pairs.
For training, we used 80,140 examples (80\% of the data) for LED, 100 examples for Llama, and 5 examples for GPT-4 5-shot.
We performed parameter and prompt tuning for all models (see Appendices \ref{apd:parameter_tuning} and \ref{apd:prompt_tuning_gpt}) and used 100 examples each for validation and testing from the remaining 20\% of the held-out data.

To evaluate lexical overlap, we used the ROUGE F1 score \citep{lin_rouge_2004}.
For similarity based on contextual embeddings, we reported BERTScore \citep{zhang_bertscore_2020} for the default \texttt{roberta-large} (BERTScore) and \texttt{microsoft/deberta-large-mnli} (DeBERT) as recommended by the authors.\footnote{See readme: \url{https://github.com/Tiiiger/bert_score}}
We did not utilize medical embeddings since the summaries should be written in layperson language.
Lastly, we determined the SARI score for text simplification \citep{xu_optimizing_2016} and the number of generated words.

\begin{figure*}[ht!]
  \centering
  \includegraphics[width=1\textwidth]{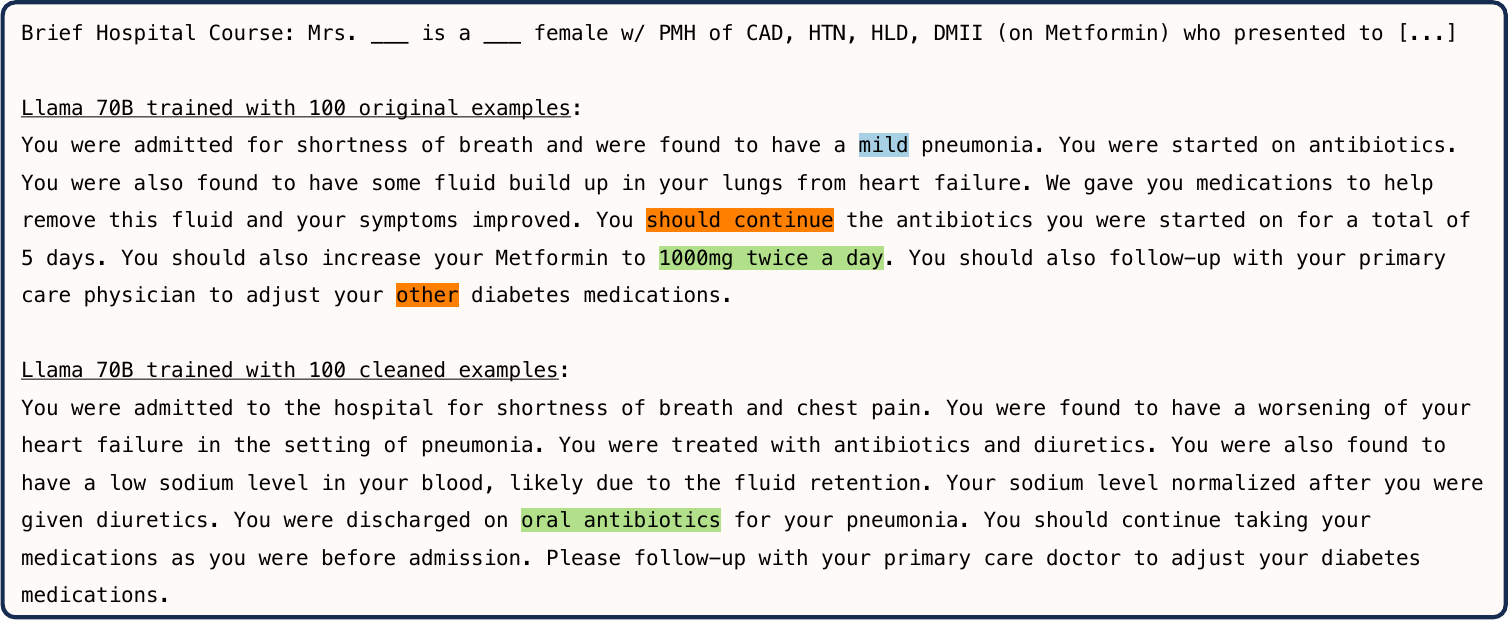}
  \caption{Patient summaries generated by Llama 70B fine-tuned on 100 \texttt{Original} and 100 \texttt{Cleaned} examples given the synthetic context in Figure \ref{fig:labelling_example} with annotated hallucinations according to our protocol. These are two of the five models included in the data-centric hallucination reduction experiments.}
  \label{fig:labelling_llama_original_cleaned}
\end{figure*}

\subsection{Qualitative Evaluation}
\label{sec:methods_qualitative_evaluation}

We evaluated the quality of patient summaries generated by LLMs aligned with the highest quality data (\texttt{Cleaned \& Improved}).
Llama 70B was fine-tuned on 100 examples and GPT-4 5-shot was prompted with 5 random examples.
We also included the original MIMIC summaries, LED-large fine-tuned on 20,942 \texttt{MIMIC-IV-Note-Ext-DI-BHC-Anno} examples, and GPT-4 0-shot (see examples in Figure \ref{fig:qualitative_examples}).
We generated summaries for the same 20 contexts from \texttt{MIMIC-IV-Note-Ext-DI-BHC-Anno} used for the hallucination-reduction experiments to reduce the annotation workload.
Ratings were performed by the same two medical experts and the summaries were again permuted to prevent model identification.
The annotators completed two training examples, and no agreement was sought as this was considered a subjective task.
We combined both ratings for our analysis.

We evaluated the summaries for \textit{Relevance}, \textit{Consistency}, \textit{Fluency}, and \textit{Coherence} on a 1-5 Likert scale \citep{fabbri_summeval_2021} following previous work on medical summarization \citep{adams_learning_2022}.
We added the dimension \textit{Simplification}.
To obtain more reliable results, we defined subtasks to be performed before rating \textit{Relevance} (subtask: labeling key facts in context and summary), \textit{Consistency} (labeling hallucinations using our protocol), and \textit{Simplification} (labeling medical jargon) and defined the meaning of each Likert rating based on these subtasks.
Additional details on the subtasks and annotator instructions are given in Appendix \ref{apd:qualitative_evaluation} and Figures \ref{fig:qualitative_evaluation_protocol} and \ref{fig:labelling_key_fact_medical_jargon}.

\subsection{Automatic Hallucination Detection}

We performed an evaluation of automated hallucination detection on two annotated datasets: 100 doctor-written summaries (\texttt{Hallucinations-MIMIC-DI}) and 100 LLM-generated summaries (\texttt{Hallucinations-Generated-DI}).
This method could serve for the automatic cleaning of training data or post-hoc editing to improve faithfulness.
We framed the task of automatic hallucination detection as a span classification problem.
Given a context $C$ and a summary $S$, our objective was to identify all spans of hallucinations within the summary, denoted as $\mathcal{H}_S = \{H_1, ..., H_n\}$.
Each hallucination span $H_i$ is defined by a tuple $(s, e, c)$, consisting of a start token, an end token, and a class (see Figure \ref{fig:labelling_example}).
We also explored class-agnostic hallucination detection without the class $c$.

We used a class-agnostic approach based on medical entities as a baseline \citep{cai_generation_2022, adams_learning_2022}.
For entity recognition and linking, we utilized MedCat \citep{kraljevic_multi-domain_2021} with UMLS semantic types from \cite{adams_whats_2021}.
All entities that appeared in the summary but not in the context were considered hallucinations.
We further enhanced this approach with SapBERT embeddings \citep{liu_self-alignment_2021} to determine equivalent entities \citep{adams_meta-evaluation_2023}.
The equivalence threshold for similarity was a hyperparameter for these methods and we determined its value based on ten additional examples labeled during annotator training.
We tested GPT-4 for class-agnostic and class-aware hallucination detection.
We designed a suitable prompt based on the aforementioned ten examples labeled during annotator training and utilized varying numbers of in-context examples and chain-of-thought (COT) prompting, detailed in Appendix~\ref{apd:hallucination_detection_gpt4}. 
Since our prompt design is based on the annotation protocol for general medical text, this approach might also prove useful for texts other than patient summaries.
Evaluation was performed using partial matching with the \texttt{nervaluate} package \citep{segura-bedmar_semeval-2013_2013}.

\begin{table*}[ht!]
\small
\setlength{\tabcolsep}{6.5pt}
\begin{tabular}{@{}lccccccccc@{}}
\toprule
\textbf{Model (training data)}                             & \textbf{R-1$\uparrow$} & \textbf{R-2$\uparrow$} & \textbf{R-3$\uparrow$} & \textbf{R-4$\uparrow$} & \textbf{R-L$\uparrow$} & \textbf{BERT$\uparrow$} & \textbf{DeBERT$\uparrow$} & \textbf{SARI$\uparrow$} & \textbf{Words} \\ \midrule
\multicolumn{10}{@{}l}{\texttt{MIMIC-IV-Note-Ext-DI-BHC} (100,175 examples)} \\ \midrule
LED-large (80,140 ex.)\hspace{0.1cm}   & $43.82$ & $17.33$ & $8.85$ & $4.92$ & $29.89$ & $88.11$ & $64.12$ & $46.71$ & $76.99$ \\
Llama 2 7B (100 ex.)  & $38.36$ & $12.66$ & $5.13$ & $2.24$ & $24.73$ & $85.68$ & $60.23$ & $44.12$ & $73.13$ \\
Llama 2 70B (100 ex.) & $40.58$ & $14.31$ & $6.09$ & $2.74$ & $26.19$ & $86.30$ & $61.89$ & $45.16$ & $76.90$ \\
GPT-4 5-shot (5 ex.) & $38.80$ & $10.78$ & $3.55$ & $1.12$ & $21.98$ & $86.67$ & $61.30$ & $42.88$ & $131.86$ \\
GPT-4 0-shot  (none)       & $38.26$ & $10.81$ & $3.70$ & $1.49$ & $21.49$ & $86.37$ & $60.75$ & $42.04$ & $165.78$ \\ 
\bottomrule
\end{tabular}
\caption{Quantitative evaluation of patient summary generation on \texttt{MIMIC-IV-Note-Ext-DI-BHC} dataset. We used the ROUGE F1-score for n-grams (R-n) and the longest common subsequence (R-L), BERTScore using \texttt{roberta-large} (BERT) and \texttt{deberta-large-mnli} (DeBERT), the SARI score, and the number of generated words. Additional performance results can be found in Table \ref{tab:performance-overview-full}.}
\label{tab:performance-overview}
\end{table*}

\section{Results}
\label{sec:results}
\subsection{Hallucinations in MIMIC and Generated Patient Summaries}
\label{sec:results_hallucination_mimic_generated}

Two medical experts labeled 100 doctor-written patient summaries (\texttt{Hallucinations-MIMIC-DI}) and found 286 hallucinations (see Table \ref{tab:label-statistics-mimic-models}).
The most prevalent label was the generic \emph{word unsupported} (n=76) followed by \emph{condition unsupported} (n=52), and \emph{time unsupported} (n=35).
Hence, there is a significant amount of unsupported data in MIMIC when using the BHC as context.
For the 100 generated patient summaries (\texttt{Hallucinations-Generated-DI}), they found a total of 114 hallucinations and \emph{word unsupported} was the most common label (see Table \ref{tab:label-statistics-mimic-models}).
Agreement statistics show that annotators agreed on 1.55 of 2.86 annotations for MIMIC and 0.67 of 1.14 annotations for the generated summaries (see Table \ref{tab:annotation-statistics}) emphasizing some variability during labeling. %
The annotation of MIMIC summaries took 30 hours for each expert and 6 hours for the agreement.
For the generated summaries, annotation took less time, with 20 hours for labeling and 4 hours for the agreement, since there were only 20 different patient contexts.

\subsection{Data-Centric Hallucination Reduction}
\label{sec:results_hallucination_reduction}

The results for GPT-4 5-shot prompted with \texttt{Original} data were significantly better than for Llama 70B fine-tuned on \texttt{Original} with 0.70 versus 2.60 hallucinations and 0.93 versus 3.77 missing key facts per summary (see Table \ref{tab:hallucination-statistics}).
Data-centric hallucination reduction (training on \texttt{Cleaned}), showed a substantial effect for Llama 70B, reducing hallucinations from 2.60 to 1.55 per summary while maintaining the same number of key facts.
Figure \ref{fig:labelling_llama_original_cleaned} displays generations of both models with hallucination annotations.
Llama 70B trained on \texttt{Original} introduced unsupported adjectives ``mild'' and ``other'', as well as an unsupported metformin dosage.
Both models incorrectly recommended continuing antibiotics at home.
Apart from ``mild'', all hallucinations can be considered reasonable advice and are commonly found in patient summaries.
For GPT-4 5-shot, the difference of 0.70 (\texttt{Original}) to 0.40 (\texttt{Cleaned}) hallucinations is only marginal, while retaining key facts.
The use of medical jargon and the length of the generations did not change substantially.
GPT-4 0-shot exhibited a few hallucinations (0.45) and the lowest number of missing key facts (0.82) and medical jargon (0.70), although the generations were longer than for 5-shot.
Table \ref{tab:label-statistics-mimic-models} shows the types of hallucination for each model.
While Llama 70B occasionally generated unsupported conditions and locations, this was rare for GPT-4.
The hallucinations produced by GPT-4 were mostly labeled as \textit{unsupported word}.

\begin{figure*}[t!]
  \hspace{-0.18cm}
  \includegraphics[width=1.016\textwidth]{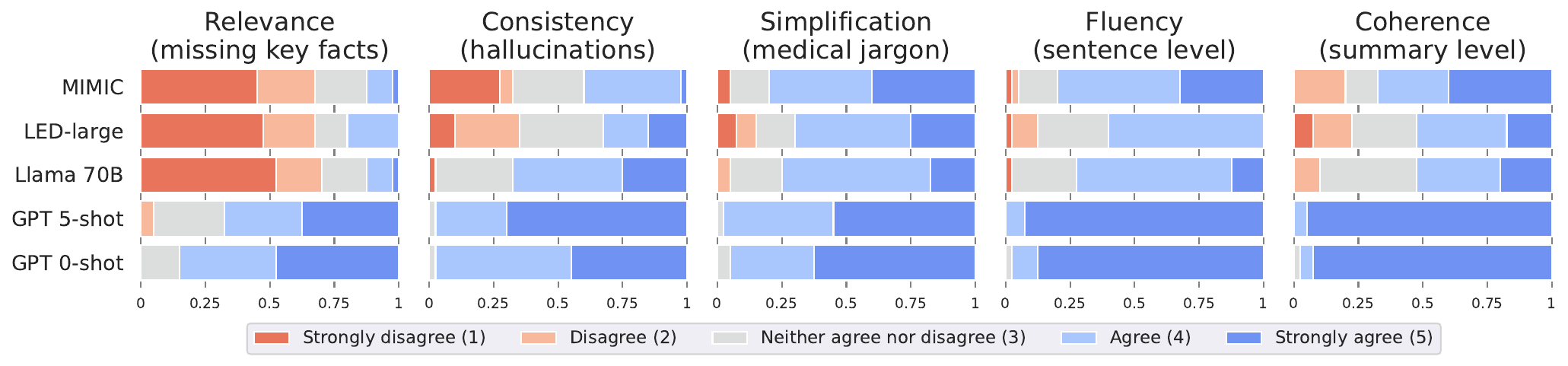}
  \caption{Qualitative evaluation of Llama 70B fine-tuned on all 100 examples of \texttt{Cleaned \& Improved} and GPT-4 5-shot prompted with 5 random examples of \texttt{Cleaned \& Improved}. We compared them to the original MIMIC summaries, LED-large fine-tuned on \texttt{MIMIC-IV-Note-Ext-DI-BHC-Anno}, and GPT-4 0-shot. Two medical experts evaluated 20 summaries from each of the five models.}
  \label{fig:qualitative_eval}
\end{figure*}

\subsection{Quantitative Evaluation}

Table \ref{tab:performance-overview} contains the performance evaluation on the \texttt{MIMIC-IV-Note-Ext-DI-BHC} dataset.
Overall, the LED model performed the best, followed by Llama and GPT-4.
Only for BERT and DeBERT, Llama and GPT-4 achieved similar performance.
This behavior is consistent with the amount of training data, which is much less for Llama and GPT-4 5-shot.
Notably, GPT-4 0-shot achieved considerable performance even without using any training data.

\subsection{Qualitative Evaluation}

The qualitative results are visualized in Figure \ref{fig:qualitative_eval}.
The \textit{Consistency} results for doctor-written (MIMIC) summaries should be interpreted with caution because these summaries were artificially considered only with the restricted context of the BHC.
The original MIMIC examples, LED-large trained on all data, and Llama 70B fine-tuned on 100 \texttt{Cleaned \& Improved} examples showed similar results.
For \textit{Relevance}, around half of the examples received a rating of two or higher, indicating that at most three key points were missing.
The \textit{Simplification}, \textit{Fluency}, and \textit{Coherence} all received a median rating of four for all three approaches.
Llama 70B performed better for \textit{Consistency} with a median rating of four compared to three for the original MIMIC and LED-large summaries.
GPT-4 performed significantly better across all five dimensions.
The difference in \textit{Consistency} compared to Llama 70B is in line with the lower hallucination rate of GPT-4 observed earlier (see Table \ref{tab:hallucination-statistics}).
Also, GPT-4 achieved high \textit{Relevance} with a median of four, indicating one missing key fact per summary, and a median of five for simplification, fluency, and coherence.
There was only a slight difference between GPT-4 5-shot and GPT-4 0-shot.
GPT-4 5-shot performed better for \textit{Consistency}, i.e., including all key facts, while GPT-4 0-shot performed better for \textit{Relevance} and \textit{Simplification}.

\begin{table}[t!]
\small
\setlength{\tabcolsep}{3.0pt}
\begin{tabular}{lrrrrrr}
\toprule
\textbf{Model}    & \multicolumn{3}{c}{\texttt{H.-MIMIC-DI}} & \multicolumn{3}{c}{\texttt{H.-Generated-DI}} \\
                  & Prec. & Rec. & F1 & Prec. & Rec. & F1 \\ 
\midrule
\multicolumn{6}{l}{Class-agnostic recognition} \\
\midrule
MedCat             & $7.4$ & $16.4$ & $10.2$ & $3.4$ & $20.3$ & $5.8$ \\
MedCat + Em. & $8.0$ & $16.1$ & $10.7$ & $3.9$ & $19.9$ & $6.5$ \\ 
GPT-4 (class-ag.)     & $23.3$ & $16.4$ & $19.3$ & $14.7$ & $16.7$ & $15.6$ \\
GPT-4 (class-aw.) & $20.0$ & $20.8$ & $20.4$ & $13.0$ & $21.5$ & $16.2$  \\
\midrule
\multicolumn{6}{l}{Class-aware recognition (11 classes)} \\
\midrule
GPT-4 (no-cot)     & $21.3$ & $4.5$ & $7.5$ & $13.4$ & $7.9$ & $9.9$   \\
\bottomrule
\end{tabular}
\caption{Results of automatic hallucination detection on 100 doctor-written (\texttt{H.-MIMIC-DI}) and 100 LLM-generated (\texttt{H.-Generated-DI}) patient summaries. GPT-4 clearly outperformed the baselines using medical UMLS entities recognized by MedCat.}
\label{tab:hallucination-recognition}
\end{table}

\subsection{Automatic Hallucination Detection}

Using medical entities extracted by MedCat for class-agnostic hallucination detection performed poorly with an F1-Score of 10.2 for doctor-written (MIMIC) and 5.8 for generated summaries (see Table \ref{tab:hallucination-recognition}). 
Adding medical embeddings (MedCat + Em.) only resulted in marginal improvements.
GPT-4 clearly outperformed this baseline on both datasets.
The best prompting strategy used a class-aware prompt (detailed in Appendix~\ref{apd:hallucination_detection_gpt4}).
GPT-4 exhibited low recall in class-aware hallucination detection highlighting the need for improved methods for automatic hallucination detection.
Additional results for class-specific recall are presented in Table \ref{tab:recall_hallucination_detection}.

\section{Discussion}
\label{sec:discussion}
Hallucinations in patient summaries are diverse and complex.
We introduced a rigorous protocol for annotating hallucinations and considered a simplified experimental setting with the BHC as limited context.
Still, it took medical experts between 12 and 18 minutes to annotate a single summary.
For the generated patient summaries, only 59\% of all annotations were identified by both annotators, with 56\% of these being of the same type (see Table \ref{tab:annotation-statistics}).
Additionally, 39\% of all hallucinations were classified as \emph{unsupported word} which typically indicates more complex hallucination structures.
The annotation of doctor-written (MIMIC) summaries yielded similar results, aligning with previous findings on consultation notes that reported medium inter-annotator agreement among clinicians \citep{moramarco_human_2022}.
We attempted to automate hallucination detection based on our annotations.
We tested UMLS concepts recognized by MedCat combined with embeddings, which showed very poor results on our datasets.
Many annotations spans did not align with medical concepts, suggesting that medical entity-based approaches may be insufficient \citep{cai_generation_2022, adams_learning_2022}.
Although GPT-4 demonstrated significantly better performance, we consider this only as an initial step. %
These results emphasize the importance of thorough evaluation with human domain experts and highlight the challenges for reliable hallucination detection in the healthcare domain.

The data used for LLM alignment is crucial for generating faithful and high-quality patient summaries.
In a study on medical summarization, \cite{adams_learning_2022} demonstrated that automatic revisions of 45,000 training examples improved the hallucination rate and quality.
In our work, we focused on more sample-efficient LLMs, where 100 human-curated training examples were sufficient.
We evaluated Llama 2 and GPT-4 for data-centric hallucination reduction, as well as for quantitative and qualitative performance.
Consistent with previous studies, we found that quantitative metrics do not correlate well with the hallucination rate \citep{moramarco_human_2022, adams_meta-evaluation_2023} and quality \citep{van_veen_adapted_2024}.
Quantitatively, LED-large outperformed both Llama 2 and GPT-4, but the ranking reversed when assessing quality, including hallucinations (\textit{Consistency}). 
This behavior can be attributed to the qualitative shortcomings of the doctor-written (MIMIC) summaries.
LED-large, which was extensively trained, most faithfully reproduced MIMIC, as confirmed by similar qualitative results.
Llama 2, fine-tuned on 100 examples, showed moderate alignment to MIMIC, placing it midway in both quantitative and qualitative evaluations.
The fine-tuning made Llama 2 more susceptible to data-centric hallucination reduction, which had a strong effect.
GPT-4 5-shot, aligned with on only five examples, largely retained its pre-trained summarization style and was less susceptible to data-centric hallucination reduction.
However, we observed that alignment with only five examples already led to slightly lower \textit{Relevance}, \textit{Simplification}, and length compared to GPT-4 without any examples (0-shot).
Hence, this study emphasizes the importance of high-quality training data for fine-tuning and in-context learning of LLMs.

The clinical evidence regarding the effectiveness of high-quality patient summaries is not clear.
\cite{becker_interventions_2021} conducted a systematic review of patient education interventions, which included 60 randomized controlled trials (RCTs).
Out of these trials, only five primarily relied on text  \citep{baker_evaluation_1991, hayes_randomized_1998, giuse_using_2012, lin_effect_2014, doyle_effect_2020}.
Closest to this work, \cite{lin_effect_2014} evaluated manually created patient-directed discharge letters based on a template, which were discussed during discharge.
They found that patient understanding improved in four domains after three and six months, but there was no significant reduction in 6-month readmission.
In this study, we focused on generating patient after-visit summaries that have only weak evidence for usefulness \citep{pathak_patient-reported_2020}.
We demonstrated that GPT-4, particularly in the 0-shot setting, deviated from the original summarization style, resulting in higher quality summaries that could potentially lead to more effective interventions \citep{haver_evaluating_2024,artsi_large_2024}.
Further enhancements could involve adaptive methods \citep{fok_qlarify_2023} to cater to different information needs and interactive approaches.
Ultimately, more clinical evidence is necessary to support the use of patient information automatically generated by LLMs.

This work has limitations.
We relied on a single dataset for medical notes and focused on patient summaries, limiting the generalizability of our findings.
We only considered the reduced context of the BHC, which likely led to an overestimation of hallucinations in doctor-written summaries.
The labeling of hallucinations was conducted by only two annotators, leading to variability in the labels.
The qualitative ratings were obtained from medical experts.
Instead, it would have been preferable to include the patients' perspectives.
Lastly, we included only two LLMs in our data-centric hallucination-reduction experiments and qualitative evaluation due to the expensive annotation required by medical experts.

\section{Conclusion}
\label{sec:conclusion}
LLMs are sensitive to fine-tuning or prompting examples.
We have demonstrated that data-centric interventions by domain experts can reduce hallucinations and improve the quality of generation.
Specifically, GPT-4 showed few hallucinations and omitted key facts while receiving high-quality ratings making it a strong model for generating patient summaries.
We have developed a protocol for labeling unsupported evidence in medical texts and have released two annotated datasets consisting of 100 doctor-written and 100 LLM-generated patient summaries.
Future work should explore different summary formats and the application of other LLMs.
The provided datasets could serve as a valuable resource for the advancement and assessment of automatic hallucination metrics.
A crucial direction for future research is the collection of more clinical evidence for useful interventions to enhance patient health literacy.

\acks{
Experiments were performed on the HPC cluster PALMA II of the University of Münster, subsidised by the DFG (INST 211/667-1).}

\bibliography{references}

\clearpage
\appendix
\section{Data Preprocessing}
\label{apd:data_preprocessing}

\subsection{Methods}
Our goal for the preprocessing was to obtain fluent patient summaries with no artifacts.
In general, we targeted for a higher precision, i.e., we considered removing low quality summaries as more important than keeping all summaries.
We used two methods to iteratively build the dataset pipeline.
First, we sampled 100 included and 100 excluded summaries and one author with medical expertise compared them with our preprocessing goals.
Based on this analysis, we repeatedly refined the dataset pipeline.
Second, we used the t-SNE method to analyze the BERT embeddings of the summaries to identify more systematic patterns.
We mostly relied on static patterns and regular expressions to remove content since automatic approaches did not show a sufficient precision.
The code for the preprocessing pipeline is available on GitHub.

\subsection{Results}
\begin{figure*}[ht!]
  \centering
  \includegraphics[width=13cm]{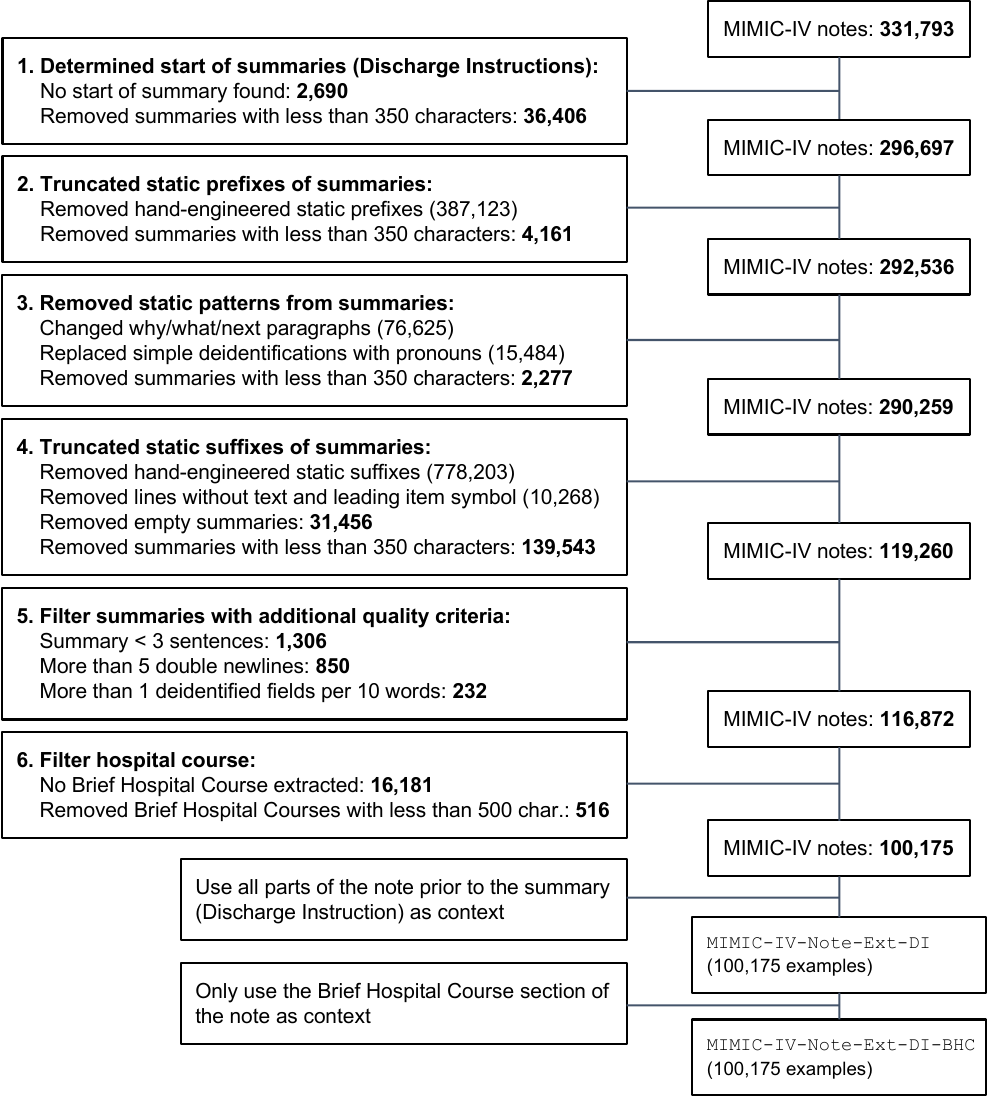}
  \caption{The preprocessing steps performed on \texttt{MIMIC-IV-Note} to obtain the datasets \texttt{MIMIC-IV-Note-Ext-DI(-BHC)}. The goal was obtain diverse and free text discharge instructions (DI) as patient summaries.}
  \label{fig:flow_chart_pipeline}
\end{figure*}

The final dataset pipeline is shown in Figure \ref{fig:flow_chart_pipeline}.
During preprocessing, we mostly focused on the summary.
After steps changing the summary length, we filtered for summaries shorter than 350 character, which we considered as useful minimum length.
To split the summary and the remaining hospital course, we simply used the \textit{Discharge Instructions} section \citep{cai_generation_2022}, which occurred in all but 2,690 notes.
This lead to a total of 296,697 candidate summaries.
In step 2, we removed static prefixes that we encountered in our analysis that contained no specific and relevant content leading to 292,536 candidates.
Next, we changed patterns in the summaries to make them more fluent.
A typical approach to structure a summary is using headings like ``Why were you in the hospital?''.
To obtain fluent and homogeneous summaries, we removed those.
Also, we replaced some simplistic deidentification patterns with the pronoun \emph{you} to reduce the deidentified content.
In step 4, we removed suffixes of the summaries, which often contained well wishes and general instructions.
We also removed several static templates, e.g., that are used after a specific surgery, by filtering for key phrases.
Hence, a large amount of summaries was removed leaving us with 119,260 notes.
Lastly, we filtered for some additional quality criteria of the summaries and the brief hospital course (BHC), which is the section before the summary leading to a total of 100,175 context-summary pairs.

We performed a performance analysis on the original \texttt{MIMIC-IV-Note} dataset and \texttt{MIMIC-IV-Note-Ext-DI} for prediction of the discharge instructions with LED-large (see Table \ref{tab:led-large-preprocessing}).
We can observe that our preprocessing pipeline lead to a significant decrease in performance, suggesting that instances that were easier to predict were removed.

\begin{table}[t!]
\small
\setlength{\tabcolsep}{4pt}
\begin{tabular}{@{}lll@{}}
\toprule
                               & \textbf{Quantity} & \textbf{Value (SD)} \\ 
\midrule
\multicolumn{3}{@{}l}{\texttt{MIMIC-IV-Note-Ext-DI} ($100,175$ context-summary ex.)} \\
\midrule
\multirow{5}{*}{Full context}  & \# Sentences      & $118.2$ ($50.4$)   \\
                               & \# Words          & $2088.8$ ($778.1$) \\
                               & \# Tokens         & $4367.1$ ($1625.3$)\\
                               & \# Characters     & $11343.6$ ($4377.8$) \\ 
                               & \# Deidentified   & $67.5$ ($39.1$)    \\
\midrule
\multirow{5}{*}{\begin{tabular}[c]{@{}l@{}}Short context (\texttt{BHC})\\\relax
[used in this work] \end{tabular}} & \# Sentences      & $33.0$ ($19.0$)  \\
                                 & \# Words          & $552.0$ ($314.0$)  \\
                                 & \# Tokens         & $858.6$ ($498.3$)\\
                                 & \# Characters     & $3029.9$ ($1736.4$)\\ 
                                 & \# Deidentified   & $11.5$ ($9.7$)    \\
                                 \midrule
\multirow{5}{*}{Summaries (DI)}  & \# Sentences      & $6.5$ ($2.6$)      \\
                                 & \# Words          & $113.2$ ($47.4$)   \\ 
                                 & \# Tokens         & $145.4$ ($61.4$)\\
                                 & \# Characters     & $604.4$ ($251.0$)  \\ 
                                 & \# Deidentified   & $1.1$ ($1.7$)    \\
\bottomrule
\end{tabular}
\caption{Overview of \texttt{MIMIC-IV-Note-Ext-DI} with the full context and \texttt{MIMIC-IV-Note-Ext-DI-BHC} with the Brief Hospital Course (BHC) as context. The discharge summaries (DI serve as patient summaries for both datasets.\protect\footnotemark}
\label{tab:dataset_overview_full}
\end{table}

\footnotetext{Sentences and words were determined with \texttt{nltk}. Tokens were determined with the Llama 2 tokenizer.}

\subsection{Analyzing 100 Included and Excluded Patient Summaries}
For a qualitative analysis of the preprocessing, we also checked 100 examples that were processed and kept in the dataset (positive) and 100 examples that were removed (negative).
One author with medical expertise looked at the original summaries and decided which part of the positive examples should be kept or whether a negative example should be removed.
Among the 100 positive examples, nine were considered problematic due to removal of important content during preprocessing.
Typically, the summary was cut at a closing statement, e.g., ``Please follow up with your PCP'', but additional content was present after this.
We also noted 33 positive examples with slight problems that did not affect the main summary.
In two examples a greetings prefix was not removed, three examples still contained parts of a template after preprocessing, and in 28 examples some fluent text was removed that was not essential for the summary (11 follow-up, 11 medication, and 6 procedure statements).
Of 100 negative examples, five were considered problematic.
All of them contained statements or headings that lead to too early suffix pruning and, hence, were removed.
Nine notes contained no or a too short brief hospital course.
Another twelve examples showed useful summaries that contained summaries between 300-350 characters.
However, we still consider this filtering useful to obtain longer summaries with more content.

\begin{figure*}[ht!]
\centering
\begin{minipage}[b]{.47\textwidth}
\hspace{-0.3cm}\includegraphics[width=1.02\textwidth]{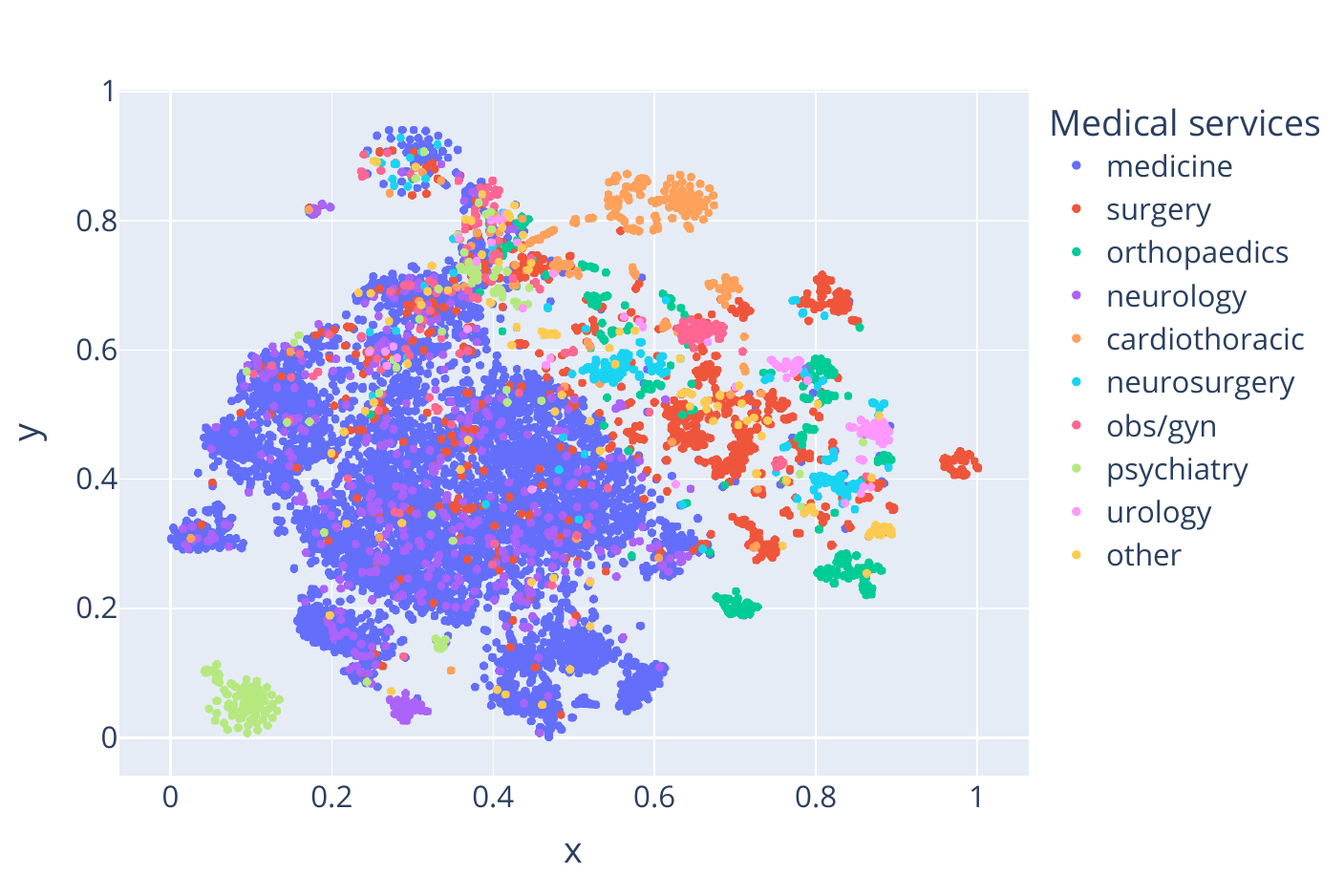}
\end{minipage}\qquad
\begin{minipage}[b]{.47\textwidth}
\includegraphics[width=1.02\textwidth]{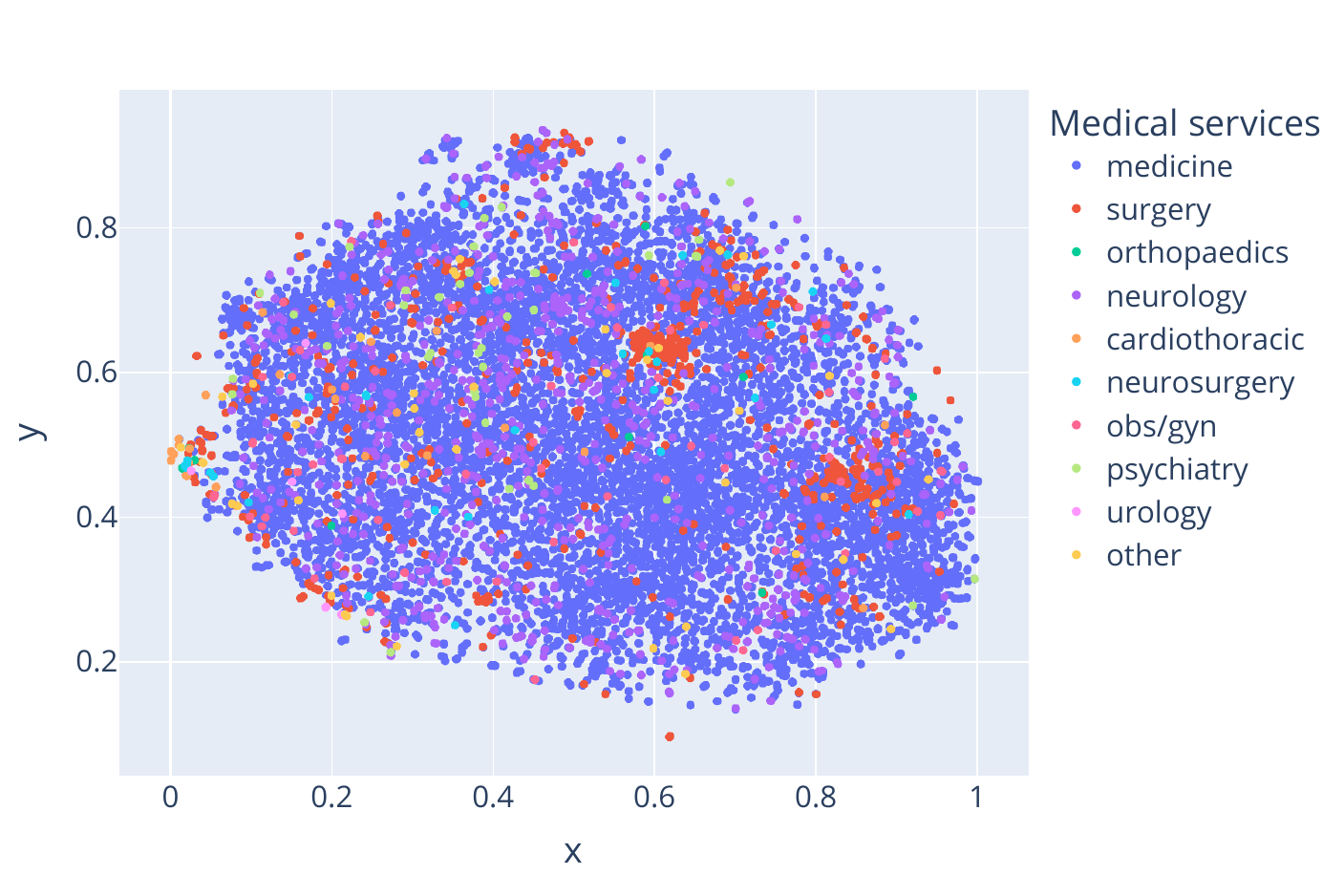}
\end{minipage}
\caption{The t-SNE embeddings of 10.000 random patient summaries before and after the preprocessing labeled with the medical specialty mentioned in the note. We can observe outliers associated with medical specialties in the unprocessed data indicating the use of summary templates.}\label{fig:embeddings_services}
\end{figure*}

\begin{table*}[ht!]
\small
\setlength{\tabcolsep}{5.5pt}
\begin{tabular}{@{}lccccccccc@{}}
\toprule
\textbf{Dataset}                             & \textbf{R-1$\uparrow$} & \textbf{R-2$\uparrow$} & \textbf{R-3$\uparrow$} & \textbf{R-4$\uparrow$} & \textbf{R-L$\uparrow$} & \textbf{BERT$\uparrow$} & \textbf{DeBERT$\uparrow$} & \textbf{SARI$\uparrow$} & \textbf{Words} \\
\midrule
\multicolumn{9}{@{}l}{LED-large (\texttt{allenai/led-large-16384})} \\
\midrule
\texttt{MIMIC-IV-Note} (unprocessed) & $51.17$ & $31.28$ & $24.06$ & $20.08$ & $39.85$ & $89.10$ & $70.48$ & $60.62$ & $162.49$ \\
\texttt{MIMIC-IV-Note-Ext-DI} & $44.04$ & $17.70$ & $9.12$ & $5.21$ & $29.89$ & $88.28$ & $64.22$ & $46.78$ & $82.15$ \\
\bottomrule
\end{tabular}
\caption{Performance results for the LED-large model before and after preprocessing. A test set of 10.000  examples was used and the remaining data for training. 
The performance of the model decreased after preprocessing, suggesting that instances that were easier to predict were removed.}
\label{tab:led-large-preprocessing}
\end{table*}

\subsection{Analyzing Embeddings}
We also inspected the t-SNE embeddings \citep{maaten_visualizing_2008} of the patient summaries before and after preprocessing.
Figure \ref{fig:embeddings_services} shows the embeddings colored by the medical service mentioned in the note.
The unprocessed summaries show several cluster that are often associated with a single medical service (same color).
Inspection of those cluster revealed that these often represent static templates and copied content.
We used these cluster to add filtering steps in our preprocessing pipeline.
The processed summaries show a much more homogeneous distribution and medical services cannot easily be distinguished.

\begin{figure*}[ht!]
  \centering
  \includegraphics[width=\textwidth]{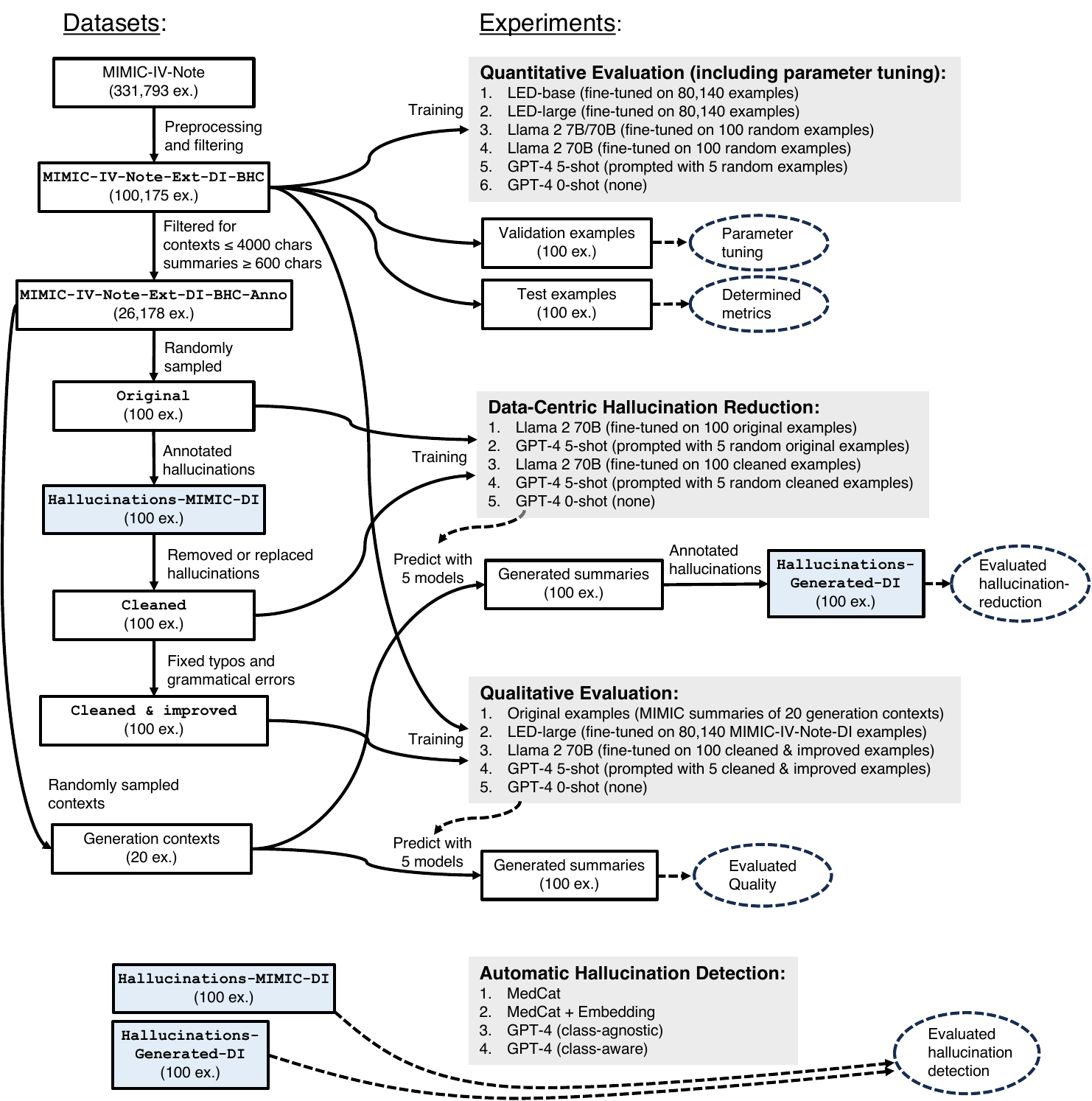}
  \caption{An overview flowchart of all datasets used in the experiments of this paper.}
  \label{fig:flow_chart_data}
\end{figure*}

\section{Annotations Statistics}
\label{apd:annotation_statistics}

Both hallucination annotations were done completely separate by both annotators.
Table \ref{tab:annotation-statistics} shows the results for MIMIC summaries and generated summaries split by annotators.
After the annotation, there was a agreement process where annotators discussed their labels and agreed on annotations.
During this process, the annotators determined the number of labels with agreement for agreement statistics.
We considered this approach more reliable than an automatic procedure since annotations of the same hallucination could differ a lot (e.g., labelling the negation instead of the subject).
Results for annotations found by both annotators with the same and different labels are also given in Table \ref{tab:annotation-statistics}.

\begin{table}[t!]
\small
\setlength{\tabcolsep}{2pt}
\begin{tabular}{lcc}
\toprule
\textbf{Quantity}    & \multicolumn{2}{c}{\textbf{Mean (SD)}} \\
                    & \texttt{H-MIMIC-DI} & \texttt{H-Generated-DI} \\ \midrule
Annotations annotator 1 & 2.39 (2.06) & 1.23 (1.52) \\
\hspace{0.4cm}Removed in agreement & 0.37 (0.66) & 0.43 (0.67) \\ \cmidrule{1-3}
Annotations annotator 2 & 2.82 (2.05) & 1.18 (1.58) \\
\hspace{0.4cm}Removed in agreement & 0.43 (0.64) & 0.17 (0.40) \\ \cmidrule{1-3}
Annotations agreement & 2.86 (2.12) & 1.14 (1.70) \\
\hspace{0.4cm}Both annot., same label & 0.99 (1.16) & 0.43 (0.78) \\
\hspace{0.4cm}Both annot., diff. label & 0.56 (0.73) & 0.24 (0.62) \\
\hspace{0.4cm}Only annotator 1 & 0.47 (0.80) & 0.13 (0.37) \\
\hspace{0.4cm}Only annotator 2 & 0.84 (1.13) & 0.34 (0.59) \\
\bottomrule
\end{tabular}
\caption{Annotation statistics for 100 doctor-written (\texttt{Hallucinations-MIMIC-DI}) and 100 generated (\texttt{Hallucinations-Generated-DI}) patient summaries. Annotation were done separately and agreed on together.}
\vspace{-0.5cm}
\label{tab:annotation-statistics}
\end{table}

We determined the inter-annotator agreement for hallucination labeling and the qualitative evaluation.
For hallucination labeling, we calculated the agreement for annotated spans using the approach from \cite{moramarco_human_2022}.
To this end, we considered a simplification of the original task and calculated the interval Krippendorff's alpha \citep{krippendorff_content_2018} for the number of annotated hallucinations per summary.
The agreement was 0.629 for the 100 MIMIC and 0.826 for the generated summaries indicating that annotating generated summaries was easier (see Table \ref{tab:inter-annotator-hallucination}).
Additionally, we computed F1-scores for the overlap of annotated spans between both annotators \citep{moramarco_human_2022}.
The results showed similar scores for MIMIC and generated summaries with F1-scores of 0.479 and 0.440 when ignoring class labels and 0.245 and 0.271 when distinguishing classes (see Table \ref{tab:inter-annotator-hallucination}).
Although the results were slightly better than those reported in \cite{moramarco_human_2022}, they underscore the difficulty of the hallucination labeling task.
To determine the inter-annotator agreement for the qualitative evaluation, we used the interval Krippendorff's alpha \citep{krippendorff_content_2018} for the Likert ratings (see Table \ref{tab:inter-annotator-qualitative}).
We distinguished the different qualitative dimensions and calculated the agreement for all ratings.
The overall agreement was 0.586, with the highest agreement observed for consistency at 0.778.
This is likely because the subtask for consistency required both annotators to label hallucinations using our protocol, which we considered the most rigorous instructions in the qualitative evaluation (see Appendix \ref{apd:qualitative_evaluation}).
The agreement for fluency and coherence was relatively low, at 0.431 and 0.218, respectively, which can be attributed to the subjective nature of these ratings.

\begin{table}[t!]
\setlength{\tabcolsep}{1.5pt}
\small
\begin{tabular}{@{}lccc@{}}
\toprule
\textbf{\begin{tabular}[l]{@{}l@{}}Annotation\\ Task\end{tabular}}     & \textbf{\begin{tabular}[c]{@{}c@{}}Agreement\\ (Kripp.-$\alpha$)\end{tabular}} & \textbf{\begin{tabular}[c]{@{}c@{}}Class-agn.\\ overlap (F1)\end{tabular}} & \textbf{\begin{tabular}[c]{@{}c@{}}Class-aw.\\ overlap (F1)\end{tabular}} \\ \midrule
MIMIC       &  $0.629$  & $0.479$ & $0.245$ \\
Generated   &  $0.826$  & $0.440$ & $0.271$ \\
\bottomrule
\end{tabular}
\caption{Inter-annotator agreement for labeling hallucinations. We determined interval Krippendorff's alpha on the number of hallucination per summary and the F1-score for overlap between annotators without (Class-agn.) and with classes (Class-aw.).}
\label{tab:inter-annotator-hallucination}
\end{table}

\begin{table}[t!]
\small
\setlength{\tabcolsep}{2.8pt}
\begin{tabular}{lcccccc}
\toprule
 & \textbf{Rel.} & \textbf{Con.} & \textbf{Sim.} & \textbf{Flu.} & \textbf{Coh.} & \textbf{Total} \\
\midrule
Agree. (Kr.-$\alpha$) & $0.457$ & $0.778$ & $0.633$ & $0.431$ & $0.218$ & $0.586$ \\
\bottomrule
\end{tabular}
\caption{Inter-annotator agreement for qualitative evaluation. We determined interval Krippendorff's alpha on the Likert ratings for each dimension separately and all ratings together  (Total).}
\label{tab:inter-annotator-qualitative}
\end{table}

\section{Parameter Tuning}
\label{apd:parameter_tuning}

\begin{table*}[t!]
\small
\setlength{\tabcolsep}{7.1pt}
\begin{tabular}{@{}lccccccccc@{}}
\toprule
\textbf{Model}                             & \textbf{R-1$\uparrow$} & \textbf{R-2$\uparrow$} & \textbf{R-3$\uparrow$} & \textbf{R-4$\uparrow$} & \textbf{R-L$\uparrow$} & \textbf{BERT$\uparrow$} & \textbf{DeBERT$\uparrow$} & \textbf{SARI$\uparrow$} & \textbf{Words} \\ \midrule
\multicolumn{10}{@{}l}{\texttt{MIMIC-IV-Note-Ext-DI-BHC} (100,175 examples)} \\ \midrule
LED-base (80,140 ex.)        & $43.32$ & $17.05$ & $8.26$ & $4.30$ & $29.21$ & $87.98$ & $63.52$ & $46.39$ & $74.36$ \\
LED-large (80,140 ex.)       & $43.82$ & $17.33$ & $8.85$ & $4.92$ & $29.89$ & $88.11$ & $64.12$ & $46.71$ & $76.99$ \\
Llama 2 7B (100 ex.)  & $38.36$ & $12.66$ & $5.13$ & $2.24$ & $24.73$ & $85.68$ & $60.23$ & $44.12$ & $73.13$ \\
Llama 2 70B (100 ex.) & $40.58$ & $14.31$ & $6.09$ & $2.74$ & $26.19$ & $86.30$ & $61.89$ & $45.16$ & $76.90$ \\
GPT-4 5-shot (5 ex.)  & $38.80$ & $10.78$ & $3.55$ & $1.12$ & $21.98$ & $86.67$ & $61.30$ & $42.88$ & $131.86$ \\
GPT-4 0-shot (none)             & $38.26$ & $10.81$ & $3.70$ & $1.49$ & $21.49$ & $86.37$ & $60.75$ & $42.04$ & $165.78$ \\ \midrule
\multicolumn{10}{@{}l}{\texttt{MIMIC-IV-Note-Ext-DI-BHC-Anno} (26,178 examples)} \\ \midrule
LED-base (20,942 ex.)        & $42.30$ & $14.98$ & $7.04$ & $3.87$ & $26.50$ & $86.71$ & $60.85$ & $44.38$ & $117.81$ \\
LED-large (20,942 ex.)       & $46.21$ & $17.38$ & $8.72$ & $5.14$ & $28.87$ & $87.50$ & $63.52$ & $45.84$ & $117.59$ \\
Llama 2 7B (100 ex.)  & $36.95$ & $11.92$ & $5.12$ & $2.53$ & $22.73$ & $82.44$ & $57.07$ & $42.41$ & $100.54$ \\
Llama 2 70B (100 ex.)  & $41.82$ & $13.63$ & $5.77$ & $2.66$ & $24.83$ & $86.43$ & $61.34$ & $43.86$ & $114.08$ \\
GPT-4 5-shot (5 ex.)              & $43.07$ & $12.91$ & $4.79$ & $2.13$ & $23.91$ & $86.80$ & $62.08$ & $43.40$ & $159.68$ \\
GPT-4 0-shot (none)             & $41.76$ & $11.39$ & $3.96$ & $1.75$ & $22.34$ & $86.42$ & $61.12$ & $42.25$ & $164.59$ \\
\bottomrule
\end{tabular}
\caption{All performance results of patient summary generation. We tested all models on the full \texttt{MIMIC-IV-Note-Ext-DI-BHC} dataset and the subset \texttt{MIMIC-IV-Note-Ext-DI-BHC-Anno} used for labeling. Parameter or prompt tuning was performed for all models as described in Appendix \ref{apd:parameter_tuning}.}
\label{tab:performance-overview-full}
\end{table*}

We performed parameter tuning for the LED \citep{beltagy_longformer_2020} and the Llama 2 \citep{touvron_llama_2023} models.
We used the full \texttt{MIMIC-IV-Note-Ext-DI-BHC} containing 100,175 examples and the subset \texttt{MIMIC-IV-Note-Ext-DI-BHC-Anno} of 26,178 examples used for labeling.
We performed a separate parameter tuning on each dataset.
For the LED model we performed full fine-tuning using 80\% of the data for training.
For Llama 2 we used LoRA for parameter-efficient fine-tuning \citep{hu_lora_2021} on 100 training examples and loaded the model in 8 bit.
For both models, we used 100 examples for validation, and 100 examples for testing.
The smaller number of validation and testing examples were chosen to reduce the runtime, however, we still consider them sufficient for representative results.
We tracked our experiments with Weights \& Biases \citep{biewald_experiment_2020}.

We used the LED models base and large (\texttt{allenai/led-\{base,large\}-16384}) from Huggingface \citep{wolf_huggingfaces_2020}.
We trained the model for 200,000 steps with a batch size of 1 and performed a validation every 20,000 steps to determine the best number of training steps.
We used a \texttt{max\_source\_length} of 4,096 and \texttt{max\_target\_length} of 350, which sufficed for almost all examples based on a prior analysis.
For the LED large model we had to used the fix described in \url{https://github.com/huggingface/transformers/issues/18190} for training.
We tuned the LED model with a complete grid search for the following parameters:
\begin{itemize}
    \item \texttt{dropout} in $\{0.05, 0.1, 0.2\}$
    \item \texttt{learning\_rate} in $\{$5e-4, 1e-5, 5e-5, 1e-6, 5e-6$\}$
\end{itemize}
We could train both models on 24 GB GPUs and the training required approximately $8$ hours for the base and $20$ hours for the large model.

For Llama 2, we also used the 7B and 70B models (\texttt{meta-llama/Llama-2-\{7,70\}b-hf} from Huggingface \citep{wolf_huggingfaces_2020}.
We trained the model for $100$ steps with a batch size of $1$ and gradient accumulation steps of $16$.
Hence, the model could encounter each training example at most $16$ times during training.
We performed a validation every $10$ steps.
The Llama 2 models have a context size of $4,096$ and we truncated the context by removing the last sentences until there were at least $350$ tokens for generation.
However, the truncation was very rarely necessary.
We tuned Llama 2 with a complete grid search for the following parameters:
\begin{itemize}
    \item \texttt{lora\_rank} in $\{8, 32\}$
    \item \texttt{lora\_alpha} in $\{8, 32\}$
    \item \texttt{lora\_dropout} in $\{0.05 0.1\}$
    \item \texttt{target\_modules} in $\{$[q\_proj, v\_proj], [q\_proj, k\_proj, v\_proj, o\_proj]$\}$
    \item \texttt{learning\_rate} in $\{$2e-5, 2e-4$\}$
\end{itemize}
For the Llama 2 7b model, we used a 24 GB GPU and the training required around 1,5 hours.
For the Llama 2 70b model, we used two 80 GB GPUs and the training required approximately 8 hours.

\section{Prompt Tuning GPT-4}
\label{apd:prompt_tuning_gpt}

\begin{table*}[t!]
\small
\setlength{\tabcolsep}{7.9pt}
\begin{tabular}{@{}lccccccccc@{}}
\toprule
GPT-4 setting & R-1 & R-2 & R-3 & R-4 & R-L & BERTScore & Deberta & SARI & Words \\ \midrule
\multicolumn{9}{@{}l}{Prompt 1: \textit{You are a helpful assistant...}} \\ \midrule
1 IC example & $40.05$ & $11.20$ & $3.87$ & $1.43$ & $19.09$ & $85.63$ & $60.53$ & $42.50$ & $208.40$ \\
3 IC examples & $40.64$ & $9.99$ & $3.51$ & $1.14$ & $19.64$ & $85.51$ & $59.94$ & $41.90$ & $189.60$ \\
5 IC examples & $39.47$ & $10.86$ & $4.32$ & $1.90$ & $20.14$ & $85.56$ & $60.56$ & $42.42$ & $198.40$ \\ \midrule
\multicolumn{9}{@{}l}{Prompt 2: \textit{You are helping with a resident working at a large urban academic medical center...}} \\ \midrule
1 IC example & $27.93$ & $5.43$ & $1.18$ & $0.05$ & $13.66$ & $83.43$ & $55.69$ & $38.54$ & $192.70$ \\
3 IC examples & $34.26$ & $7.34$ & $2.50$ & $1.06$ & $17.90$ & $85.26$ & $59.58$ & $40.44$ & $131.80$ \\
5 IC examples & $32.92$ & $7.28$ & $2.41$ & $0.63$ & $17.40$ & $85.12$ & $59.33$ & $38.65$ & $127.70$ \\ \midrule
\multicolumn{9}{@{}l}{Prompt 3: \textit{You are a helpful assistant that helps patients understand their medical records...}} \\ \midrule
0 IC examples           & $42.50$ & $11.95$ & $4.37$ & $2.09$ & $21.49$ & $86.30$ & $61.36$ & $45.70$ & $214.40$ \\
1 IC examples & $36.05$ & $8.97$ & $3.28$ & $1.53$ & $18.32$ & $85.84$ & $60.20$ & $43.10$ & $191.80$ \\
3 IC examples & $38.54$ & $9.71$ & $3.33$ & $1.35$ & $19.16$ & $85.72$ & $60.55$ & $43.25$ & $199.00$ \\
5 IC examples & $39.22$ & $10.73$ & $3.95$ & $1.69$ & $20.73$ & $85.65$ & $60.06$ & $42.39$ & $206.30$ \\
5 IC examples + ``You...'' & $41.99$ & $12.83$ & $5.22$ & $2.26$ & $22.67$ & $86.95$ & $62.35$ & $43.55$ & $138.70$ \\\bottomrule
\end{tabular}
\caption{Performance results for GPT-4 for three different prompt formats with different numbers of in-context (IC) examples. The results were generated on ten held-out summaries.}
\label{tab:gpt4-prompt-performance}
\end{table*}

We performed prompt tuning for GPT-4 for generating patient summaries.
We tried three different prompt formats (detailed in Table~\ref{tab:gpt_summarization_prompts}) and evaluated them qualitatively and quantitatively on ten held-out examples.
The prompts used different settings of a general assistant, a medical resident assistant, and a patient assistant.
We noticed that the results of the second prompt contained a lot of medical jargon leading to results targeted at medical experts.
The results for the first prompt contained some simplifications but did not follow the style of the in-context examples.
For instance, often a different start was used ``During your hospitalization...''.
Prompt three led to generations with simplified language that followed the style of the in-context examples.
Hence, qualitatively we considered the third prompt format the best.
We tried two additional variations of prompt three using no in-context examples and using five in-context examples plus an instruction to start with ``You were admitted'', which should help to guide the generation format.
The second variant further improved the qualitative results.

This finding was supported by the quantitative analysis on ten held-out examples from the MIMIC-IV-Note-Ext-DI dataset.
The results are shown in table \ref{tab:gpt4-prompt-performance}.
In general, prompt three lead to the best results and additional in-context example did not degrade the performance.
Guiding the output with the instructions to start with ``You were admitted'' also lead to a quantitative improvement.
Hence, we chose this prompt for all our GPT-4 summarization experiments.

\section{Hallucination Detection with GPT-4}
\label{apd:hallucination_detection_gpt4}

\begin{table*}[ht!]
\small
\setlength{\tabcolsep}{4.65pt}
\begin{tabular}{@{}lccccccccccc@{}}
\toprule
\textbf{Model}               & cond. & proc. & medic. & time & location & number & name & words & other & contrad. & incorr. \\ \midrule
\multicolumn{12}{@{}l}{\texttt{Hallucinations-MIMIC-DI}} \\ \midrule
MedCat             & $28.8$ & $31.6$ & $22.1$ & $2.9$ & $6.9$ & $7.1$ & $2.8$ & $15.8$ & $50.0$ & $13.3$ & $0.0$ \\
MedCat + Embedding & $28.8$ & $31.6$ & $22.1$ & $2.9$ & $5.2$ & $7.1$ & $2.8$ & $15.1$ & $50.0$ & $13.3$ & $0.0$ \\
GPT-4 (class-ag.)      & $9.6$ & $18.4$ & $27.9$ & $17.1$ & $17.2$ & $50.0$ & $8.3$ & $7.2$ & $50.0$ & $46.7$ & $0.0$           \\ 
GPT-4 (class-aw.)     & $14.4$ & $26.3$ & $33.8$ & $24.3$ & $19.0$ & $57.1$ & $11.1$ & $10.5$ & $50.0$ & $46.7$ & $0.0$           \\ \midrule
\multicolumn{12}{@{}l}{\texttt{Hallucinations-Generated-DI}} \\ \midrule
MedCat             & $38.5$ & $31.8$ & $30.8$ & $0.0$ & $30.8$ & $0.0$ & $0.0$ & $6.4$ & $0.0$ & $25.0$ & $0.0$ \\
MedCat + Embedding & $38.5$ & $31.8$ & $26.9$ & $0.0$ & $30.8$ & $0.0$ & $0.0$ & $6.4$ & $0.0$ & $25.0$ & $0.0$ \\
GPT-4 (class-ag.)      & $11.1$ & $62.5$ & $25.0$ & $50.0$ & $0.0$ & $83.3$ & $40.0$ & $5.7$ & $0.0$ & $42.9$ & $0.0$           \\ 
GPT-4 (class-aw.)     & $18.5$ & $50.0$ & $45.0$ & $50.0$ & $0.0$ & $83.3$ & $50.0$ & $10.2$ & $0.0$ & $35.7$ & $0.0$           \\ \bottomrule
\end{tabular}
\caption{Recall for different hallucination labels for hallucination detection on 100 doctor-written summaries (\texttt{Hallucinations-MIMIC-DI}) and 100 generated summaries (\texttt{Hallucinations-Generated-DI}) using partial matching.}
\label{tab:recall_hallucination_detection}
\end{table*}

Our GPT-4 hallucination detection pipeline consists two steps: (1) prompt GPT-4 to annotate hallucinated spans in the original summary text and (2) identify and extract the annotated span. 

As GPT-4 is a decoder-only model, we cannot directly uses it to produce per-token predictions for potential spans of hallucination.
One approach is to prompt it to ``label'' hallucination spans in the input text, i.e., to generate an HTML-like tag \styledtext{oc-orange-9}{oc-orange-0}{\texttt{\textless{}error class="hallucination type"\textgreater{}}} to indicate potential spans of hallucination in context. 
For example, for a sentence in the summary text, ``Your pacemaker rate was increased to 50'', GPT-4 would generate the following annotated sentence: ``Your pacemaker rate was increased to \styledtext{oc-orange-9}{oc-orange-0}{\texttt{\textless{}error class="unsupported\_number"\textgreater{}}}50\styledtext{oc-orange-9}{oc-orange-0}{\texttt{\textless{}/error\textgreater{}}}'' when the actual number is 40. 
This approach is similar to generate the relation or unverifiable labels in the pipeline by \citet{mishra_fine-grained_2024}.

Given the generated annotations, we then extract the labeled spans from the annotated text and use them to identify potential hallucinations.
Sometimes GPT-4 might produce text that is slightly different from the original text (e.g., correcting typos in the original sentences).
Under such circumstances, we first match the generations with the original sentence and then extract the labeled spans.

Table \ref{tab:gpt_hallucination_detection_prompts1},\ref{tab:gpt_hallucination_detection_prompts2},\ref{tab:gpt_hallucination_detection_prompts3} show the prompt used to inform GPT-4 of potential error types and the format for annotating the hallucination spans. 
We can optionally turn off the hallucination class detection (i.e., generating only \styledtext{oc-orange-9}{oc-orange-0}{\texttt{\textless{}error\textgreater{}}} rather than \styledtext{oc-orange-9}{oc-orange-0}{\texttt{\textless{}error class="hallucination type"\textgreater{}}}), and we compare the results as class-aware and class-agnostic hallucination detection. 

One can also optionally prompt the LLM to reason about potential spans of hallucination before annotating the full text with chain-of-thought reasoning.
In practice, we prompt GPT-4 to generate a list of labeled hallucination and explanations that mimics the examples in the prompt, i.e., generating bullet point lists like \texttt{- "Your <error>red blood cell count</error> was followed and was stable." The BHC does not state that the red blood cell count was followed. Instead the hematocrit remained stable according to the BHC.} with explanation before annotating the full summary with hallucination labels.

\begin{figure*}
\centering
\fbox{
\begin{minipage}[t][19.6cm][t]{0.97\textwidth}
\begin{enumerate}
    \item \textbf{Relevance}: The rating measures how well the summary captures the key points of the brief hospital course. Consider whether all and only the important aspects are contained in the summary. \textit{To this end, please label key facts in the context and the summary.}\\\\
    5 = all key points included\\
    4 = 1 key point missing\\
    3 = 2 key point missing\\
    2 = 3 key point missing\\
    1 = more than 3 key points missing
    
    \item \textbf{Consistency}: The rating measures whether the facts in the summary are consistent with the facts in the original brief hospital course. Consider whether the summary does reproduce all facts accurately and does not make up untrue information. \textit{To this end, please label all hallucination in the summary according to the protocol.}\\\\
    5 = no finding\\
    4 = 1-2 minor findings\\
    3 = more than 2 minor or 1 major finding\\
    2 = 2 major findings\\
    1 = more than 2 major findings
    
    \item \textbf{Simplification}: The rating measures whether the summary is written in plain language understandable for a patient. Consider medical terms and abbreviations. \textit{To this end, please label medical jargon in the summary.}\\\\
    5 = no medical jargon\\
    4 = 1-2 minor unexplained medical term \\
    3 = more than 2 minor or 1 major medical term\\
    2 = 2 major medical terms\\
    1 = more than 2 major medical terms
    
    \item \textbf{Fluency}: This rating measures the quality of individual sentences, are they well-written and grammatically correct. Consider the quality of individual sentences.\\\\
    5 = all sentences are well-written and correct\\
    4 = 1-2 sentences have minor errors or poor style\\
    3 = more than 2 sentences with minor errors or poor style or 1 sentence with severe error\\
    2 = 2 sentences with major errors\\
    1 = more than 3 major errors

    \item \textbf{Coherence}: The rating measures the quality of all sentences collectively, to the fit together and sound naturally. Consider the quality of the summary as a whole.\\\\
    5 = the summary as a whole is well-written and clearly structured\\
    4 = the summary is mostly well-written and structured\\
    3 = the summary can be followed and has some structure\\
    2 = the summary is hard to follow and has no clear structure\\
    1 = the summary is very hard to follow and very unstructured
\end{enumerate}
\end{minipage}}
\caption{Instructions for qualitative evaluation. For step 1, 2, and 3 additional annotation were performed.}
\label{fig:qualitative_evaluation_protocol}
\end{figure*}

\section{Qualitative Evaluation}
\label{apd:qualitative_evaluation}

For our qualitative evaluation of patient summaries we used the dimension introduced in \cite{fabbri_summeval_2021}.
They evaluated the quality of summaries for news articles in the CNN/DailyMail dataset \citep{hermann_teaching_2015}.
The same taxonomy was already used for medical summaries \citep{adams_learning_2022}.
We adapted the instructions slightly by replacing ``article`` with ``brief hospital course`` to better suite our setting.
We also added the dimension \textit{Simplification} to measure the extent to which layperson language was used.
We presented the annotators with the instructions shown in Figure \ref{fig:qualitative_evaluation_protocol}.
For step 2, we required the annotators to label hallucination.
We used the developed protocol for this.
For steps 1 and 3, annotators had to label key facts and medical jargon as described below.

\subsection{Labeling Key Facts and Medical Jargon}

\begin{figure*}[t!]
  \centering
  \includegraphics[width=\textwidth]{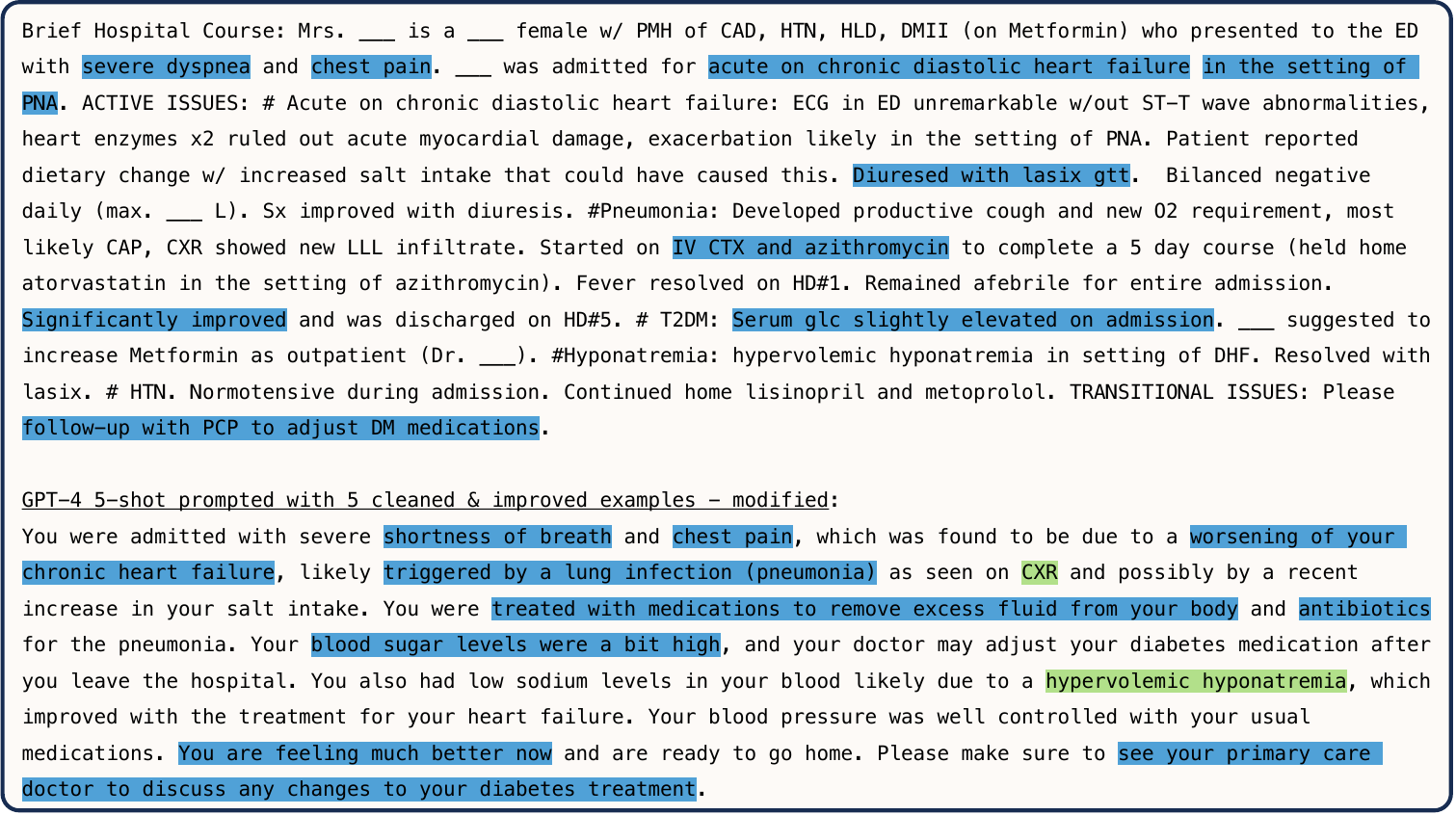}\vspace{0.1cm}
    \caption{Example for annotation of key facts and medical jargon based on the synthetic example and a modified output of GPT-4 5-shot. Annotators labeled key facts in the context and the summary (dark blue). The number of missing facts was determined based on the difference between key facts in the context and the respective summary. Medical jargon was labeled only in the summary (green). In the given summary, there are zero missing key facts and two uses of medical jargon.}
    \label{fig:labelling_key_fact_medical_jargon}
\end{figure*}

To improve the quality of the qualitative evaluation, we required annotators to label key facts and medical jargon for step 1 and 3.
Based on this, they would then enter their rating on a Likert scale.
We also used annotation for key facts and medical jargon to ensure the quality of generated summaries in our hallucination-reduction experiments (see Table \ref{tab:hallucination-statistics}).

We did not develop a protocol for these labeling tasks.
Instead, we used a more subjective procedure.
For labeling key facts, annotators were asked to label key information that is important for the patient in the context.
After that, they should label mentions of these facts in each summary (see dark blue annotations in Figure \ref{fig:labelling_example}).
Based on this, missing key facts in the summary could be determined.
For medical jargon, annotators were asked to label span that they assumed would be difficult to be understood by the given patient (see green annotations in Figure \ref{fig:labelling_example}).
Every medical jargon term was only counted once for each summary.
We also did not perform an agreement procedure between both annotators, since we did not develop a formal protocol and considered these labeling tasks more subjective.

\begin{figure*}[t!]
  \centering
  \includegraphics[width=\textwidth]{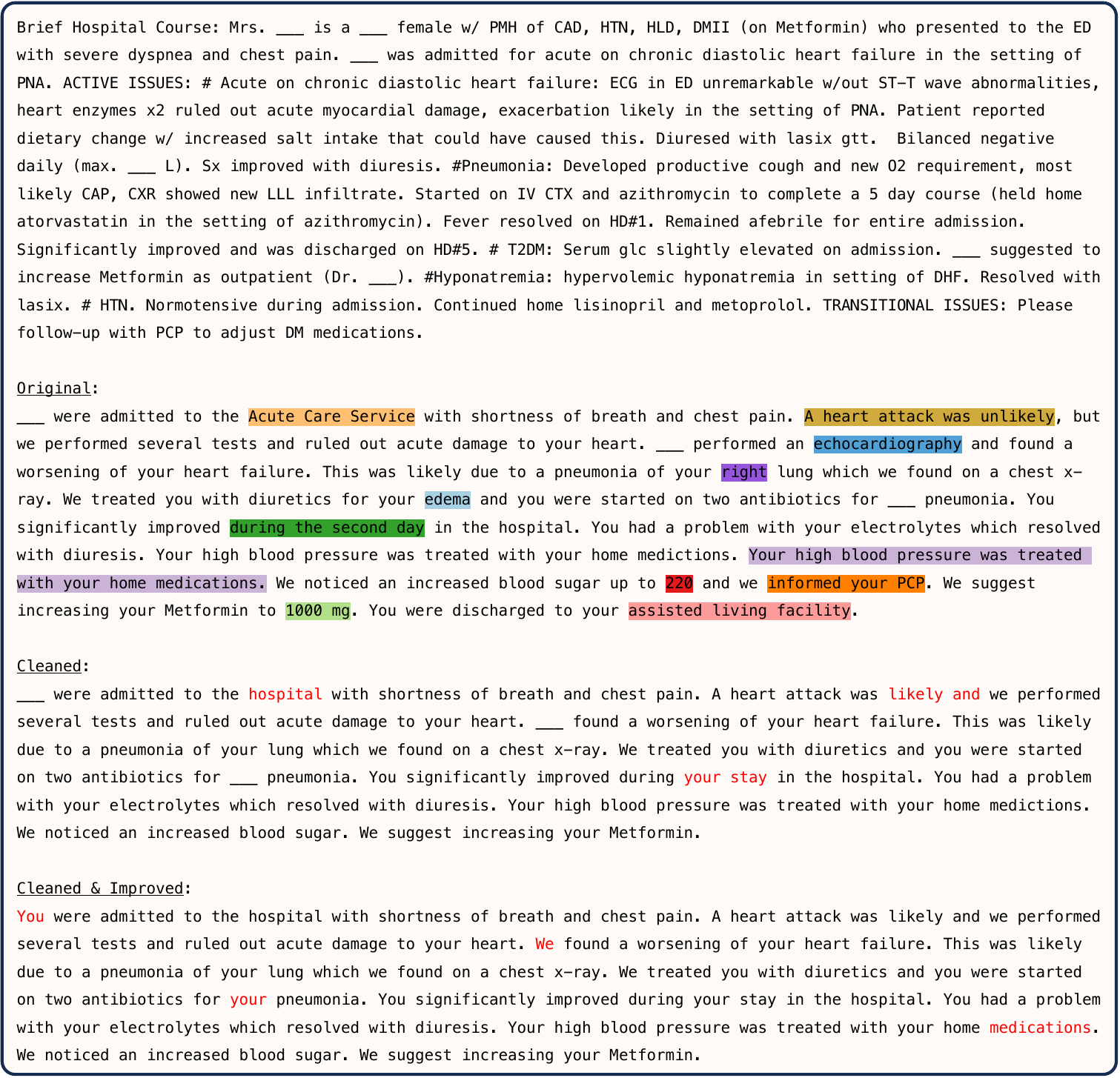}\vspace{0.1cm}
  \vspace{-0.7cm}
  \caption{Example for the creation of the derived datasets \texttt{Cleaned} and \texttt{Cleaned \& Improved}. The example is a slightly modified version of Figure \ref{fig:labelling_example}. For \texttt{Cleaned}, we manually replaced hallucinations with useful facts in the context or removed them otherwise. For \texttt{Cleaned \& Improved}, we further corrected mistakes and artifacts such as typos or deidentifications. We performed this process for all 100 examples in the \texttt{Original} dataset.}
  \label{fig:derived_datasets}
  \vspace{-0.5cm}
\end{figure*}

\begin{figure*}[t!]
  \centering
  \includegraphics[width=\textwidth]{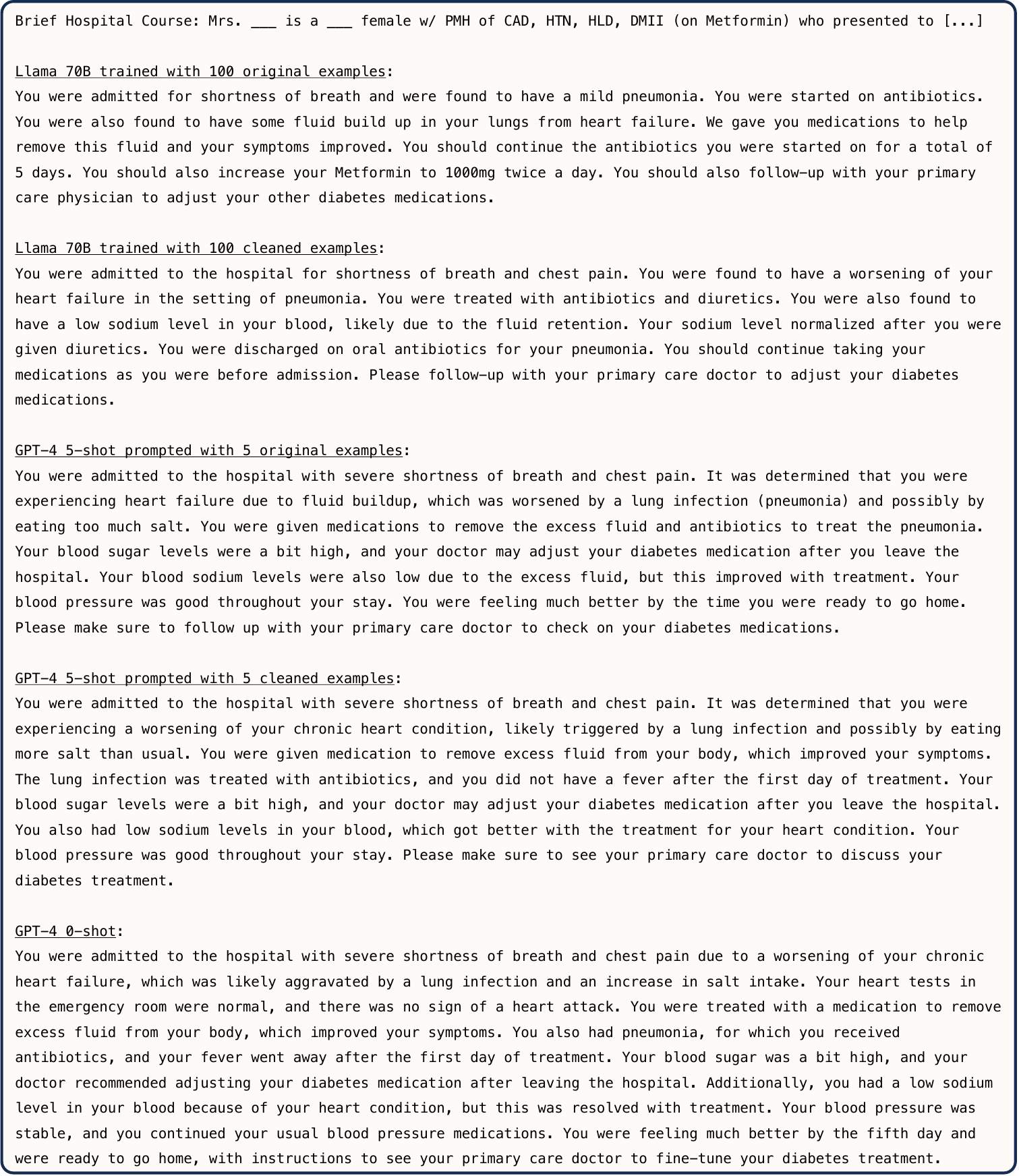}\vspace{0.1cm}
  \vspace{-0.7cm}
  \caption{Examples for the data-centric hallucination reduction. We generated a summary with all models included in the hallucination reduction experiment given the synthetic BHC as context (see Figure \ref{fig:labelling_example}). Examples for Llama 70B are identical to Figure \ref{fig:labelling_llama_original_cleaned}.}
  \label{fig:hallucination_examples}
  \vspace{-0.5cm}
\end{figure*}

\begin{figure*}[t!]
  \centering
  \includegraphics[width=\textwidth]{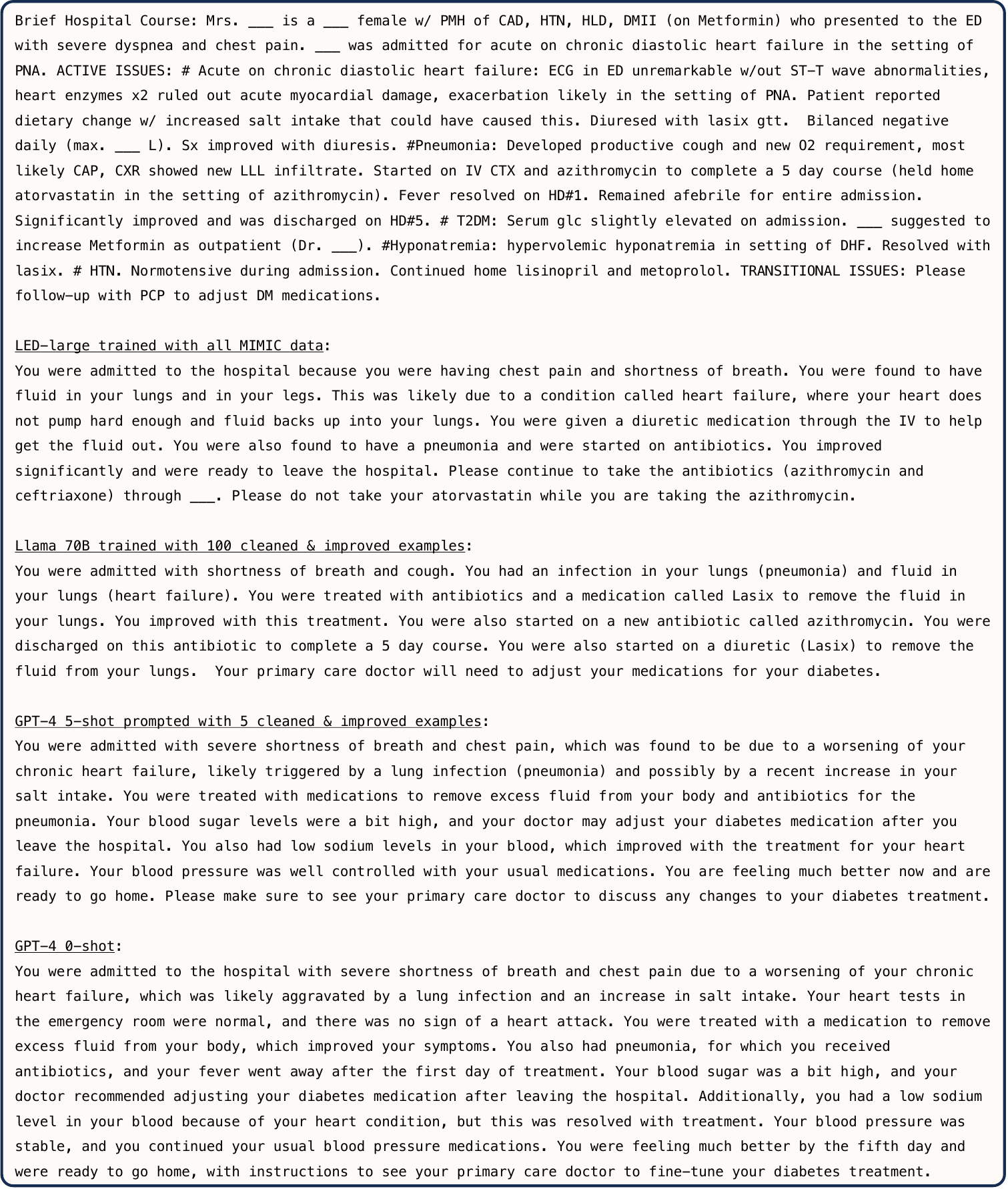}\vspace{0.1cm}
  \vspace{-0.7cm}
  \caption{Examples for the qualitative evaluation. We generated a summary with all models included in the qualitative evaluation given the synthetic BHC as context (see Figure \ref{fig:labelling_example}).}
  \label{fig:qualitative_examples}
  \vspace{-0.5cm}
\end{figure*}

\begin{table*}[t]
  \resizebox{1.\linewidth}{!}{
  \small
  \centering
  \setlength{\tabcolsep}{3.5pt}
  \begin{tabular}{lm{0.9\linewidth}}
  \multicolumn{2}{c}{\textsc{Prompt 1}} \\ 
  \midrule
  \textbf{System} & 
  \begin{ttblock}
  You are a helpful assistant.
  \end{ttblock}
  \\
  \midrule
  \textbf{User} & 
  \begin{ttblock}
  You will be given a doctor's note and you will need to summarize the patient's brief hospital course.

  Let's do a practice round.\newline 
  \{\{\#each icl\_examples\}\}\newline 
  Here is the doctor's note on a patient's brief hospital course:\newline
  \prompttemplate{\{\{this.document\}\}}\newline
  Summarize for the patient what happened during the hospital stay based on this doctor's note. Please make it short and concise and only include key events and findings. \newline
  \prompttemplate{\{\{this.summary\}\}}\newline
  \{\{/each\}\}
  \newline
  Here is the doctor's note on a patient's brief hospital course:\newline
  \prompttemplate{\{\{test\_document\}\}}\newline
  Summarize for the patient what happened during the hospital stay based on this doctor's note. Please make it short and concise and only include key events and findings. %
  \end{ttblock}
  \\
  \midrule
  \textbf{Assistant} & 
  \begin{ttblock}
  \promptgeneration{\{\{generate\_summary (max\_tokens=600 temperature=0)\}\}}
  \end{ttblock} 
  \\
  \midrule
  \\
  \multicolumn{2}{c}{\textsc{Prompt 2}} \\ 
  \midrule
  \textbf{System} & 
  \begin{ttblock}
  You are helping with a resident working at a large urban academic medical center.
  \end{ttblock}
  \\
  \midrule
  \textbf{User} & 
  \begin{ttblock}
  You task is to help summarize a patient's brief hospital course based on the doctor's note. Please make it short and concise and only include key events and findings. \newline
  Here are some examples:\newline
  \{\{\#each icl\_examples\}\}\newline
  DOCUMENT: \newline
  \prompttemplate{\{\{this.document\}\}}\newline
  SUMMARY: \newline
  \prompttemplate{\{\{this.summary\}\}}\newline
  \{\{/each\}\}\newline
  Here is another doctor note on a patient's brief hospital course:\newline
  DOCUMENT: \prompttemplate{\{\{test\_document\}\}}
  \end{ttblock}
  \\
  \midrule
  \textbf{Assistant} & 
  \begin{ttblock}
  \promptgeneration{\{\{generate\_summary (max\_tokens=600 temperature=0)\}\}}
  \end{ttblock} 
  \\
  \midrule
  \\
  \multicolumn{2}{c}{\textsc{Prompt 3}} \\ 
  \midrule
  \textbf{System} & 
  \begin{ttblock}
  You are a helpful assistant that helps patients understand their medical records. 
  \end{ttblock}
  \\
  \midrule
  \textbf{User} & 
  \begin{ttblock}
  You will be given some doctor's notes and you will need to summarize the patient's brief hospital course in one paragraph. Please only include key events and findings and avoid using medical jargons, and you MUST start the summary with "You were admitted". \newline %
  Here are some examples:
  \{\{\#foreach icl\_examples\}\}\newline
  DOCUMENT: \newline
  \prompttemplate{\{\{this.document\}\}}\newline
  SUMMARY: \newline
  \prompttemplate{\{\{this.summary\}\}}\newline
  \{\{/each\}\}\newline
  DOCUMENT: \prompttemplate{\{\{test\_document\}\}}
  \end{ttblock}
  \\
  \midrule
  \textbf{Assistant} & 
  \begin{ttblock}
  \promptgeneration{\{\{generate\_summary (max\_tokens=600 temperature=0)\}\}}%
  \end{ttblock} 
  \\ 
  \bottomrule
  \end{tabular}
  }
\caption{Different prompts for using GPT-4 to generate patient summaries. In all examples, we can use a list of in-context learning examples (\texttt{icl\_examples}, whether cleaned or not) to guide the models for the final generation of the target summary, allowing 600 new tokens as well as using greedy decoding.}
\label{tab:gpt_summarization_prompts}
\end{table*}

\begin{table*}[t]
  \resizebox{1.\linewidth}{!}{
  \small
  \centering
  \renewcommand{\arraystretch}{1.15}
  \setlength{\tabcolsep}{3.5pt}
  \begin{tabular}{m{1.0\linewidth}}
  \toprule
  \begin{ttblock}
  We will present you with a pair of a brief hospital course (BHC) and a patient after visit summary (AVS). The AVS is also referred to as discharge summary. The BHC contains a detailed summary of the hospital stay written by medical service. It usually contains medical jargon, and it can follow different structures based on the hospital course and responsible medical specialty. The AVS summarizes the hospital stay for the patient in plain language. In practice, the BHC is not the only source of information to write the AVS. However, in our setting we treat the BHC as the only context for the summary.
  \newline \newline   
  \#\# Instructions 
  \newline \newline  
  For this labelling task, we are interested in errors in the AVS that are either unsupported by the BHC, contradict content in the BHC, or are wrong medical facts. We allow statements that contain general medical knowledge or advice that are often used in patient summaries. Most errors are due to unsupported facts, so we further distinguish those based on their specific content. This leads to the following error types or labels:\newline
  1. Unsupported facts, including condition/procedure/medication/time/location/\newline number/name/word/other\newline
  2. Contradicted fact\newline
  3. Incorrect fact\newline
  And below is the detailed guideline, and we label error spans with the \textless{}error\textgreater{} tag (e.g. \textless{}error class="error\_type"\textgreater{}incorrect fact\textless{}/error\textgreater{}).
  \newline\newline
  \#\#\# Determining Span of Errors\newline
  We label the smallest possible consecutive span that specifies the error given the BHC as a context. Removing further parts from the span would remove important information. A useful heuristic is to identify the minimal span that must be replaced to obtain a correct statement that is grammatically correct. For example\newline
  - "We performed an \textless{}error\textgreater{}esophageal-gastro-duodenoscopy (EGD).\textless{}error\textgreater{}" when no such procedure is reported in the BHC. The article "an" is not labeled as an error. When no procedure at all was performed "performed an esophageal-gastro-duodenoscopy (EGD)" should be labeled as error because there is no suitable substitute for "esophageal-gastro-duodenoscopy (EGD)".\newline
  - "After the surgery, we \textless{}error\textgreater{}transitioned you to oral oxycodone\textless{}/error\textgreater{}." when the BHC contains no information for such a transition. If another medication transition is mentioned in the BHC and makes sense in this sentence only "oral oxycodone" should be labeled. If another oral medication transition is mentioned in the BHC only "oxycodone" should be labeled.\newline
  - "\textless{}error\textgreater{}Your symptoms responded well\textless{}/error\textgreater{}." when no part of the sentence makes sense in the given context of the AVS.
  \newline\newline  
  We allow general medical knowledge and advice that is often part of the AVS. Usually, these are information that are not specific for the hospital course given in the BHC. For example\newline
  - "Please take your medications as prescribed" contains no error even though the BHC does not contain this instruction because this is general medical advice.\newline
  - "If the symptoms get worse, please contact your doctor" contains no error even when the BHC does not contain this fact, since it is general medical knowledge that a doctor should be seen for worsening symptoms. \newline
  \newline
  We try to ignore grammatical errors in the BHC and AVS. If the original meaning can still be inferred (e.g. "medictaions" instead of "medications"), the most likely corrected form can be used. If the meaning cannot be inferred, they can be ignored in the BHC or labeled as Unsupported Other in the AVS.\newline
  \newline
  If a sentence or phrase is repeated, then please treat it as you would any other sentence and highlight all errors (even if you did so in a previous sentence). For example\newline
  - "Please take Tylenol. Please take Tylenol" when Tylenol was prescribed in the BHC.\newline
  - "Limit your \textless{}error\textgreater{}use of stairs\textless{}/error\textgreater{}. Please limit \textless{}error\textgreater{}use of stairs\textless{}/error\textgreater{}" when movement was encouraged.
  \end{ttblock}
  \\
  \bottomrule
  \end{tabular}
  }
\caption{Prompts for using GPT-4 to detect hallucinations (Part 1). We only show the user message as the system prompt is the same as in Table \ref{tab:gpt_summarization_prompts}.}
\label{tab:gpt_hallucination_detection_prompts1}
\end{table*}

\begin{table*}[t]
  \resizebox{1.\linewidth}{!}{
  \small
  \centering
  \renewcommand{\arraystretch}{1.15}
  \setlength{\tabcolsep}{3.5pt}
  \begin{tabular}{m{1.0\linewidth}}
  \toprule
  \emph{Continued from Table~\ref{tab:gpt_hallucination_detection_prompts1}} \newline\newline
  \begin{ttblock}
  To get reliable error counts a span should only contain a single error.\newline
  - "You received \textless{}error\textgreater{}Tylenol\textless{}/error\textgreater{} and \textless{}error\textgreater{}Ciprofloxacin\textless{}/error\textgreater{}" when there is no evidence in the BHC that the two medications were administered to the patient.\newline
  - "You have a \textless{}error\textgreater{}follow-up appointment with your PCP\textless{}/error\textgreater{} and \textless{}error\textgreater{}your cardiologist\textless{}/error\textgreater{}" when no such follow up is mentioned in the BHC. Both errors are labeled separately.\newline
  \newline
  \#\#\# Dealing with Deidentified Information\newline
  The data contains deidentified information shown with "\_\_\_" in the text. We always treat this as non-existent information. So, the annotators should not infer what the deidentified information could be. In general, deidentified fields in the AVS should not be labeled as errors. However, sometimes they belong to a wrong statement or clearly contain unsupported information (e.g., a doctor's name or phone numbers) that are not given in the BHC. In these cases, deidentified fields should be included in the error span. For example\newline
  - "Take \_\_\_ \textless{}error\textgreater{}200mg daily\textless{}/error\textgreater{} and try to rest" when no such dosage information is provided in the BHC, but the statement to rest. The deidentified medication name is excluded from the error span.\newline
  - "Please avoid going up \textless{}error\textgreater{}more than \_\_\_ stairs\textless{}/error\textgreater{} at a time" when restrictions for the number of stairs taken at a time are note mentioned in the BHC.\newline
  - "\textless{}error\textgreater{}Dr. \_\_\_ will follow up with you\textless{}/error\textgreater{}" when no follow-up is mentioned in the BHC.\newline
  - "Please stop taking Aspirin \textless{}error\textgreater{}on \_\_\_\textless{}/error\textgreater{}" when no stopping date is given in the BHC. \newline
  - "Your RBC peaked \textless{}error\textgreater{}at \_\_\_ million\textless{}/error\textgreater{}" if there is no hint of a specific red blood cell count given in the BHC.\newline
  \newline
  \#\#\# Error Types\newline
  In general, we ask for the most specific error that is applicable. If there is uncertainty which type applies, prefer the one mentioned first in the enumeration of all error types shown earlier. For instance, if the error contains an unsupported medication name, the Unsupported medication type should be used instead of the Unsupported name type. Here is a detailed description of the error types:\newline
  - `Unsupported Condition`: includes unsupported symptoms, diseases, or findings of the patient. For example\newline
  \hspace*{2em}- "You were found to have a \textless{}error class="unsupported\_condition"\textgreater{}left clavicle fracture\textless{}/error\textgreater{}" when no information was given for this condition in the BHC.\newline
  - `Unsupported Procedure`: includes any unsupported medical procedures. For example\newline
  \hspace*{2em}- "You had a \textless{}error class="unsupported\_procedure"\textgreater{}filter placed in your vein\textless{}/error\textgreater{}" when no intervention with a filter was mentioned.\newline
  - `Unsupported Medication`: contains all errors related to unsupported medications. This includes medication classes, substances, routes, frequencies, and dosages. For example\newline
  \hspace*{2em}- "You were placed on \textless{}error class="unsupported\_medication"\textgreater{}antibiotics\textless{}/error\textgreater{}" when only blood thinners were prescribed.\newline
  - `Unsupported Time`: includes all errors for unsupported time or interval statements. For example\newline
  \hspace*{2em}- "Keep your arm in a sling for the \textless{}error class="unsupported\_time"\textgreater{}next 6 weeks\textless{}/error\textgreater{}" when no specific duration is given.\newline
  - `Unsupported Location`: Locations include both unsupported physical places as well as regions of the patient. For example\newline
  \hspace*{2em}- "The patient was admitted to the \textless{}error class="unsupported\_location"\textgreater{}Acute Surgery Service\textless{}/error\textgreater{}" when no admission location was provided in the BHC.\newline
  - `Unsupported Number`: any number either as digits or written that are unsupported. This also includes words such as "a" and "an". For example\newline
  \hspace*{2em}- "Your pacemaker rate was increased to \textless{}error class="unsupported\_number"\textgreater{}50\textless{}/error\textgreater{}" when the rate of 50 is not given in the BHC.
  \end{ttblock}
  \\
  \bottomrule
  \end{tabular}
  }
\caption{Prompts for using GPT-4 to detect hallucinations (Part 2). }
\label{tab:gpt_hallucination_detection_prompts2}
\end{table*}

\begin{table*}[t]
  \resizebox{1.\linewidth}{!}{
  \small
  \centering
  \renewcommand{\arraystretch}{1.15}
  \setlength{\tabcolsep}{3.5pt}
  \begin{tabular}{m{1.0\linewidth}}
  \toprule
  \emph{Continued from Table~\ref{tab:gpt_hallucination_detection_prompts2}} \newline\newline
  \begin{ttblock}
  - `Unsupported Name`: named entities that are not supported by the BHC. For example\newline
  \hspace*{2em}- "You were seen by the \textless{}error class="unsupported\_name"\textgreater{}interventional pulmonary service\textless{}/error\textgreater{}" when no consult with this service was mentioned in the BHC.
  - `Unsupported Word`: incorrect or inappropriate words or phrases which do not fit in any of the above types. For example\newline
  \hspace*{2em}- "We will send you home with a \textless{}error class="unsupported\_word"\textgreater{}drain\textless{}/error\textgreater{} in place" when drain not mentioned in the BHC.\newline
  - `Unsupported Other`: If there is a mistake which clearly does not belong to any of the above categories, you may use this category as a last resort. We cannot give precise instructions because the "other" category is very broad.\newline
  - `Contradicted Fact`: This error type is independent of the content and contains all facts that clearly contradict information provided in the BHC. For example\newline
  \hspace*{2em}- "Your pacemaker rate was increased to \textless{}error class="contradicted\_fact"\textgreater{}50\textless{}/error\textgreater{}" when the context state a pacemaker rate of 40.\newline
  - `Incorrect Fact`: This error type is independent of the content and contains all facts that clearly contradict general medical knowledge or advice. For example\newline
  \hspace*{2em}- "We diagnosed a seizure, and you \textless{}error class="incorrect\_fact"\textgreater{}can continue driving your car\textless{}/error\textgreater{}" when no reason for allowing driving after a seizure is provided this contradict common medical knowledge.\newline
  \newline
  \#\# Examples\newline
  \newline
  {\{\{\#each icl\_examples\}\}}\newline
  {\#\#\# Example \prompttemplate{\{\{this.index\}\}}}\newline
  \newline
  BHC: \newline
  \prompttemplate{\{\{this.document\}\}}\newline
  \newline
  AVS: \newline
  \prompttemplate{\{\{this.summary\}\}}\newline
  \newline
  ERRORS:\newline
  \prompttemplate{\{\{this.cot\_description\}\}}\newline
  \newline
  AVS WITH ERRORS LABELED:\newline
  \prompttemplate{\{\{this.summary\_with\_errors\}\}}\newline
  \{\{/each\}\}\newline
  \newline
  \#\#\# Example \prompttemplate{\{\{n\_shot+1\}\}}\newline
  \newline
  BHC: \newline
  \prompttemplate{\{\{test\_document\}\}}\newline
  \newline
  AVS: \newline
  \prompttemplate{\{\{test\_summary\}\}}\newline
  \newline
  ERROR:
  \end{ttblock}
  \\
  \bottomrule
  \end{tabular}
  }
\caption{Prompts for using GPT-4 to detect hallucinations (Part 3). }
\label{tab:gpt_hallucination_detection_prompts3}
\end{table*}

\clearpage
\let\clearpage\relax


\section*{Supplementary material}

used. If the meaning cannot be inferred, they can be ignored in the BHC or labeled as *Unsupported Other* in the AVS.

### Repeated Content
If a sentence or phrase is repeated, then please treat it as you would any other sentence and highlight all errors (even if you did so in a previous sentence). For example
- "Please take Tylenol. Please take Tylenol" when Tylenol was prescribed in the BHC.
- "Limit your <u>use of stairs</u>. Please limit <u>use of stairs</u>" when movement was encouraged.

### Complex Mistakes
If there are multiple ways in which you can annotate a sentence, follow the following rules:
1. Choose the alternative with fewest total mistakes.
2. Choose the alternative with the smallest error spans.
3. If both annotations would lead to the same number of mistakes with the same span sizes, prefer the error mentioned first in the enumeration of all error types above.

## Additional Clarifications

### Clarifications from first batch of 5 test examples
- If a condition is a consequence of another condition, it is not labeled as an error even if it is not directly mentioned.
  - For example, "You had pain during your admission" if a severe trauma is mentioned in the BHC.
- If another doctor takes over the treatment, it can be assumed that they re-evaluate the patient and can decide the following treatment.
  - For example, "Please continue your medications until your PCP appointment" if no stopping time for the medications was mentioned in the BHC.
- Facts classified as general medical knowledge are considered from a doctor's perspective. The doctors write the AVS; hence, this general knowledge needs no support in the BHC but can be assumed to be known by any doctor.
  - For example, "After we placed a drug-eluting stent you will receive two blood thinning medications" if the BHC mentions the placement if a drug-eluting stent, the antiplatelet therapy can be assumed as general knowledge.
- If a medical specialty commonly treats a condition or performs a procedure, this is considered general knowledge and not considered as an unsupported fact.
  - For example, "Neurology evaluated your EEG" if the EEG evaluation was mentioned in the BHC but though no specialty was given.

### Clarifications from second batch of 5 test examples
- Specific names of departments are not treated as general knowledge even though it is likely that they treat a condition or perform a procedure. We require those to be explicitly named in the BHC.

- o For example, "Neuro Epilepsy Service evaluated your EEG" if the EEG evaluation was mentioned in the BHC but no responsible department was given.
- The main condition for admission named in the summary ("You were admitted for …") cannot only be derived from the main condition named in the BHC. Sometimes another major reason relevant to the patient is mentioned later in the BHC, so it is also acceptable to use this.
  - o For example, "You were admitted for abdominal pain." If the BHC states an ileus as the main reason but also mentions the abdominal pain.

## Clarifications from agreement procedures of 100 examples

- A reference to additional material or information that are not part of the AVS (e.g., medication plan or phone number) are considered as unsupported facts.
  - o For example, "Your medication plan is attached." If no medication plan is mentioned in the BHC.
- Generalizations and simplifications of medical conditions or procedures that are still close to the information in the BHC and aid patient understanding are not considered as unsupported facts.
  - o For example, "You WBC was low" when the BHC mentions a neutropenia.
- If a span contains different error types, it is labeled as "unsupported word".
  - o For example, "The lab test showed an increased risk of bleeding." when no lab test or increased risk of bleeding is mentioned in BHC. The error is labeled with one span since logically they belong together.
- If a common daily dosage of a medication is given, we treat instructions for daily intake in the AVS as general knowledge.
  - o For example, "You therapy was changed to Plavix 75 mg daily." If the BHC only contains the therapy change to Plavix 75 mg.
- If unsupported entities are clearly referred to in the next sentence, they are again labeled as an error.
  - o For example, "Vascular Surgery was consulted. They recommended no surgery." if the vascular surgery team is not mentioned in the BHC.


\end{document}